\documentclass[10pt,twocolumn,letterpaper]{article}

\usepackage{overpic}
\usepackage{adjustbox}
\usepackage{tikz}
\usepackage{xcolor}
\usepackage{colortbl}
\RequirePackage{graphicx}
\RequirePackage{amsmath}
\RequirePackage{amssymb}
\RequirePackage{algorithm}
\RequirePackage{algpseudocode}
\usepackage{booktabs}
\definecolor{cvprblue}{rgb}{0.21,0.49,0.74}
\usepackage[pagebackref,breaklinks,colorlinks,allcolors=cvprblue]{hyperref}
\definecolor{deepgreen}{RGB}{0, 190, 0}

\usepackage[pagenumbers]{iccv} 

\newcommand{\tophline}{\specialrule{1.5pt}{0pt}{0pt}}
\newcommand{\midhline}{\specialrule{0.75pt}{0pt}{0pt}}
\newcommand{\bottomline}{\specialrule{1.5pt}{0pt}{0pt}}
\usepackage{url}
\usepackage{times}
\usepackage{epsfig}
\usepackage{graphicx}
\usepackage{amsmath}
\usepackage{amssymb}
\usepackage{multirow}
\usepackage{cases}
\usepackage{booktabs}
\usepackage{color}
\usepackage{cases}
\usepackage{wasysym}
\usepackage{comment}

\definecolor{iccvblue}{rgb}{0.21,0.49,0.74}

\def\paperID{*****} 
\def\confName{ICCV}
\def\confYear{2025}

\title{Local Dense Logit Relations for Enhanced Knowledge Distillation}

\author{
	Liuchi Xu\textsuperscript{\rm 1} \quad  Kang Liu\textsuperscript{\rm 2} \quad  Jinshuai Liu\textsuperscript{\rm 1} \quad Lu Wang\textsuperscript{\rm 1}\footnotemark[1] \quad Lisheng Xu\textsuperscript{\rm 1} \quad Jun Cheng\textsuperscript{\rm 3,4}\thanks{Corresponding authors}\\ 
 \textsuperscript{\rm 1} Northeastern University, China;
 \textsuperscript{\rm 2} South China Normal University, China;\\
\textsuperscript{\rm 3}
Shenzhen  Institutes of Advanced Technology, CAS, China;\\
 \textsuperscript{\rm 4}The Chinese University of Hong Kong, Hong Kong, China.\\
 \{xuliuchi@stumail,liujinshuai@stumail,wanglu@mail\}.neu.edu.cn,\\	2022010207@m.scnu.edu.cn,xuls@bmie.neu.edu.cn,Jun.cheng@siat.ac.cn
}

\begin{document}
\maketitle
\begin{abstract}
\noindent State-of-the-art logit distillation methods exhibit versatility, simplicity, and efficiency.~Despite the advances, existing studies have yet to delve thoroughly into fine-grained relationships within logit knowledge.~In this paper, we propose Local Dense Relational Logit Distillation (LDRLD), a novel method that captures inter-class relationships through recursively decoupling and recombining logit information, thereby providing more detailed and clearer insights for student learning.~To further optimize the performance,~we introduce an Adaptive Decay Weight (ADW) strategy, which can dynamically adjust the weights for critical category pairs using Inverse Rank Weighting (IRW) and Exponential Rank Decay (ERD).~Specifically, IRW assigns weights inversely proportional to the rank differences between pairs, while ERD adaptively 
controls weight decay based on total ranking scores of category pairs.~Furthermore, after the recursive decoupling, we distill the remaining non-target knowledge to ensure knowledge completeness and enhance performance.~Ultimately, our method improves the student's performance by transferring fine-grained knowledge and emphasizing the most critical relationships.~Extensive experiments on datasets such as CIFAR-100, ImageNet-1K, and Tiny-ImageNet demonstrate that our method compares favorably with state-of-the-art logit-based distillation approaches. The code will be made publicly available.
\end{abstract}    
\section{Introduction}
\label{sec:intro}

Deep Convolutional Neural Networks (CNNs) have been widely applied in various computer vision tasks, including image classification~\cite{he2016deep,yang2022cross}, object detection~\cite{zheng2023localization,zhao2024detrs}, and semantic segmentation~\cite{zou2024segment,liang2023open}.~To enhance performance, researchers have designed increasingly complex networks~\cite{woo2023convnext} that demand greater computational and memory resources. However, deploying these high-performance, parameter-heavy networks on resource-constrained devices, such as smartphones and cameras~\cite{gou2021knowledge}, remains challenging.~In response to this issue, researchers have proposed several techniques to decrease the model size and computational cost, such as efficient network design~\cite{zhang2018shufflenet,howard2017mobilenets,sandler2018mobilenetv2}, weight pruning~\cite{anonymous2024infobatch}, low-rank factorization~\cite{haeffele2019structured}, quantization~\cite{duan2023qarv}, and knowledge distillation (KD)~\cite{hinton2015distilling}.~Among these techniques, KD is particularly efficient and practical. It facilitates the development of lightweight networks suitable for real-time mobile applications without altering the original network architecture.
\begin{figure}[!t]
\center{\includegraphics[width=8.3cm]{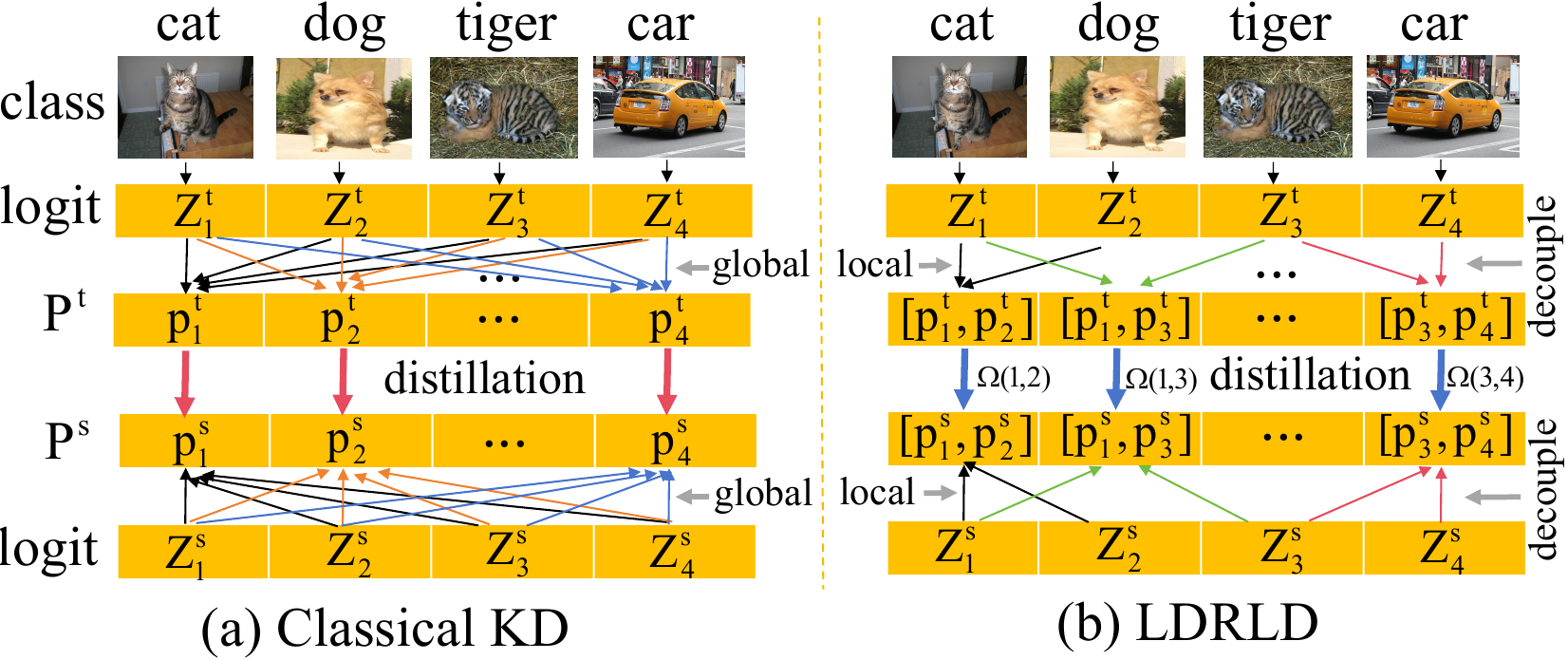}} 
\caption{~(a) Hinton et al.~\cite{hinton2015distilling} introduce KD through global softmax, which calculates the probability, leading to information redundancy between classes and diminishing the logit discrimination. In classical KD, the prediction probability difference between ``cat" and ``dog" derived from teacher's logit output is calculated as: $\Delta P_{KD}=|p_{1}^{t}-p_{2}^{t}|$=$|\frac{\exp({\mathbf{Z_{1}^{t}}})-\exp({\mathbf{Z_{2}^{t}}})}{\sum_{i=1}^C{\exp({\mathbf{Z_{i}^{t}}})}}|$. (b) In contrast, our proposed LDRLD uses category pairs and calculates probability difference between ``cat" and ``dog" as: $\Delta P_{LDRLD}=|p_{1}^{t}-p_{2}^{t}|$=$|\frac{\exp({\mathbf{Z_{1}^{t}}})-\exp({\mathbf{Z_{2}^{t}}})}{\sum_{i=1}^2{\exp({\mathbf{Z_{i}^{t}}})}}|$. It is obvious that $\Delta P_{LDRLD}>\Delta P_{KD}$, indicating that our approach enhances inter-class differences 
compared to KD and improves fine-grained logit discrimination.}
\label{kd and ldrld}
\vspace{-14pt}
\end{figure}

The core idea of KD is often to transfer the ``dark knowledge" from a larger pre-trained teacher model (teacher) to a more compact student model (student) at the logit (or output) layer, providing higher-level semantic information.~Generally, KD uses the Kullback-Leibler ($\mathcal{KL}$) divergence~\cite{kullback1951information} with a temperature coefficient to minimize the difference between the teacher's and student's output probability distributions.~KD methods are commonly categorized into two main types:~feature-based KD~\cite{romero2014fitnets,komodakis2017paying,xiaolong2023norm,zongbetter,miles2024understanding,shen2024student} and logit-based KD~\cite{zhao2022decoupled,yang2023knowledge,luo2024scale,xu2024improving,zheng2024knowledge,li2022asymmetric,jin2023multi}.~Feature-based KD methods aim to align knowledge derived from intermediate layers, which contain rich feature representations.~Despite their effectiveness, feature-based KD methods demand considerable computational and memory resources, thus posing significant training challenges.~Consequently, researchers have shifted their focus to logit-based KD methods to reduce resource consumption.~These methods often decouple logit knowledge~\cite{zhao2022decoupled,yang2023knowledge,luo2024scale,xu2024improving,li2022asymmetric,KDExplainer} or employ dynamic temperature techniques~\cite{Sun2024Logit,jin2023multi,li2023curriculum,zheng2024knowledge} to extract critical information more effectively, thus boosting performance, computational efficiency, and versatility.

Considering these advancements, it is important to note that while current logit-based KD methods have achieved impressive performance, they may fail to adequately capture fine-grained inter-class relationships and 
enhance inter-class distinction, resluting in the limitations specified as follows.~Since global softmax  focuses on high-probability classes~\cite{hinton2015distilling}, it may reduce the differences between lower-probability classes, thereby limiting the student's ability to capture their fine-grained logit relationships. Moreover, intuitively, categories with greater semantic differences are generally easier to distinguish. However, classical KD may introduce redundant information from other classes when modeling the two-class relationships. As shown in Fig.~\ref{kd and ldrld} (a), the distinction between ``cat" and ``dog" is reflected through probability distributions, yet the introduction of irrelevant (e.g., ``car") can interfere with and weaken their inter-class discriminability.~These limitations collectively hinder the student's ability to capture unique category characteristics, potentially degrading its performance. Therefore, it is essential to enhance inter-class discriminability while comprehensively leveraging logit relationships.
 
In this paper, we propose a Local Dense Relational Logit Distillation (LDRLD) method, which recursively decouples and recombines logit knowledge to provide richer information for student learning.~By virture of its effective capture of fine-grained logit relationships, LDRLD can enhance the efficacy of the distillation process. Based on this capability, we observe that categories with closer semantic relationships are more challenging to distinguish, whereas more distant categories are easier to recognize due to their semantic dissimilarities. To further refine our approach and enhance its performance, we introduce an Adaptive Decay Weight (ADW) strategy that dynamically adjusts weights based on ranked category pairs in LDRLD.~By employing Inverse Rank Weighting (IRW) and Exponential Rank Decay (ERD), ADW assigns larger weights to closely related category pairs and smaller weights to distant ones.~Specifically, IRW assigns larger weights to pairs with smaller rank differences and smaller weights to those with larger differences, whereas ERD dynamically adjusts weight decay based on the sum of the rankings.~This dynamic weighting scheme enhances the student's ability to optimize performance on visually complex and critical categories. Furthermore, after the recursive decoupling, we also distill the remaining non-target knowledge~\cite{zhao2022decoupled} to ensure completeness. By integrating the ADW strategy into LDRLD, we optimize the utilization of the logit knowledge, which leads to better model decision boundaries.~Because of this optimization, the student can capture inter-class relationships and obtain rich contextual knowledge.~By enhancing the student's ability to identify critical categories, the proposed method effectively mitigates classification ambiguities and ensures accurate knowledge transfer.

~To validate the effectiveness of our method, we conduct comprehensive experiments across various datasets.~The contributions of this paper are summarized as follows:

\begin{itemize}
  \item We propose Local Dense Relational Logit Distillation (LDRLD), a novel method that captures fine-grained logit relationships more effectively and enhances inter-class  discriminability.

  \item We introduce the Adaptive Decay Weight (ADW) strategy, which comprises Inverse Rank Weighting (IRW) and Exponential Rank Decay (ERD). ADW can dynamically adjust the weights of ranked category pairs, thus enabling the student to more effectively optimize the classification of challenging categories.
  \item Extensive experimental results on diverse datasets, including CIFAR-100, ImageNet-1K, and Tiny-ImageNet, consistently show that our method outperforms existing state-of-the-art logit-based KD methods, and justify its ability to capture and transfer critical inter-class relationships.

\end{itemize}


\section{Related Work}
In this section, we will review existing works that relate to feature-based and logit-based knowledge distillation (KD).
\subsection{Feature-based Knowledge Distillation}
Extensive works~\cite{yang2022cross,yang2022focal,zhao2024detrs,romero2014fitnets,komodakis2017paying,tian2019contrastive,chen2021distilling,liu2023functionconsistent,guo2023class,huang2024knowledge,kim2024do,yang2022masked,yue2020matching,yang2021knowledge,wang2019distilling} have shown that the intermediate features contain rich and diverse information, which helps students in acquiring and mastering knowledge more effectively.~Specifically, FitNets~\cite{romero2014fitnets} uses mean squared error (MSE) between the teacher's and student's intermediate features to guide the student training.~AT~\cite{komodakis2017paying} adopts activation maps as attention mechanisms to replicate the teacher's intermediate features more accurately.~VID~\cite{ahn2019variational} boosts information transfer between the teacher and the student by increasing the mutual information (MI) of intermediate features.~CRD~\cite{tian2019contrastive} proposes using contrastive learning for knowledge distillation, where contrastive targets are employed to enhance the transfer of the teacher’s representations to the student.~ReviewKD~\cite{chen2021distilling} treats shallow knowledge as old knowledge and guides the student's learning through repeated review.~FCFD~\cite{liu2023functionconsistent} optimizes the functional similarity between teacher and student features, thereby improving the student's mimicry efficiency.~CAT-KD~\cite{guo2023class} facilitates the student's learning via the transfer of the teacher's class activation maps. Although feature-based KD methods provide rich information, their high computational and memory demands have shifted research towards using logit knowledge as a more efficient distillation objective.
\subsection{Logit-based Knowledge Distillation} 
\textbf{Understanding the Role of Knowledge Distillation.} Early research explores the underlying mechanisms of KD, for the purpose of seeking deeper insights~\cite{zhang2018deep,mirzadeh2020improved,zhang2022quantifying,furlanello2018born,yang2019snapshot,tang2020understanding,zhou2021rethinking,stanton2021does,muller2019does,shen2021label,phuong2019towards,cho2019efficacy,bayesian,ye2024bayes,zhao2023dot,zhou2021wsl}. For instance, Tang et al.~\cite{tang2020understanding} theoretically demonstrate the potential of the label smoothing (LS) technique for improving student performance.~Muller et al.~\cite{muller2019does} discover that LS can reduce intra-class variations and result in more compact feature representations.~Furthermore, Shen et al.~\cite{shen2021label} show that tight clustering enhances the separability of similar semantic representations across categories.
Despite these advancements, the performance of vanilla KD methods, such as deep mutual learning~\cite{zhang2018deep}, teacher assistants~\cite{mirzadeh2020improved,son2021densely}, BAN~\cite{furlanello2018born}, and early stopping~\cite{cho2019efficacy}, may still fall short of practical application demands.

\noindent\textbf{Refining Logit Knowledge for Improving Distillation Performance.} Recent studies have achieved better performance by focusing on dynamic temperature and decoupling the logit knowledge to streamline distillation.

\textit{1) Dynamic temperature.}~Vanilla KD methods use a fixed temperature, which can limit the effectiveness of transferring logit knowledge.~To overcome this limitation, dynamic temperature techniques have been proposed to improve the efficiency of logit knowledge transfer~\cite{Sun2024Logit,jin2023multi,li2023curriculum,zheng2024knowledge}.~For example, CTKD~\cite{li2023curriculum} adopts an adaptable temperature to modulate task difficulty.~MLKD~\cite{jin2023multi} employs Gram matrix distillation with varying temperatures to reduce prediction biases.~LSKD~\cite{Sun2024Logit} proposes adjusting the temperature based on the weighted standard deviation of logits and seamlessly integrates Zero-Score normalization.~WTTM~\cite{zheng2024knowledge} uses transformed teacher matching to highlight temperature scaling in knowledge refinement while omitting it on the student side.

\textit{2) Decoupling the logit knowledge.}
Recent studies have focused on decoupling logit knowledge to develop more effective KD methods~\cite{zhao2022decoupled,yang2023knowledge,li2022asymmetric,xu2024improving,luo2024scale,Auxiliary}.~For example,
ATS~\cite{li2022asymmetric} enhances performance by decomposing the $\mathcal{KL}$ divergence into correction guidance, label smoothing, and class distinguishability.~DKD~\cite{zhao2022decoupled} proposes decoupling the $\mathcal{KL}$ divergence into target
and non-target class
knowledge distillation, better exploiting information to improve performance.~NKD~\cite{yang2023knowledge} presents normalizing non-target logits to enhance the utilization of soft labels in distillation. 
ReKD~\cite{xu2024improving} divides logit knowledge into head and tail categories to facilitate knowledge transfer.~SDD~\cite{luo2024scale} decouples global and local logit outputs into consistent and complementary terms to enhance the student’s ability to acquire precise logit knowledge.~These methods improve student's performance by absorbing knowledge more efficiently.


\section{Method}
In this section, we first review the vanilla knowledge distillation (KD) and then introduce our proposed Local Dense Relational Logit Distillation (LDRLD) method.~After that, we present the optimization objective of the LDRLD.
\begin{figure*}[!ht]
\center{\includegraphics[width=17.0cm]{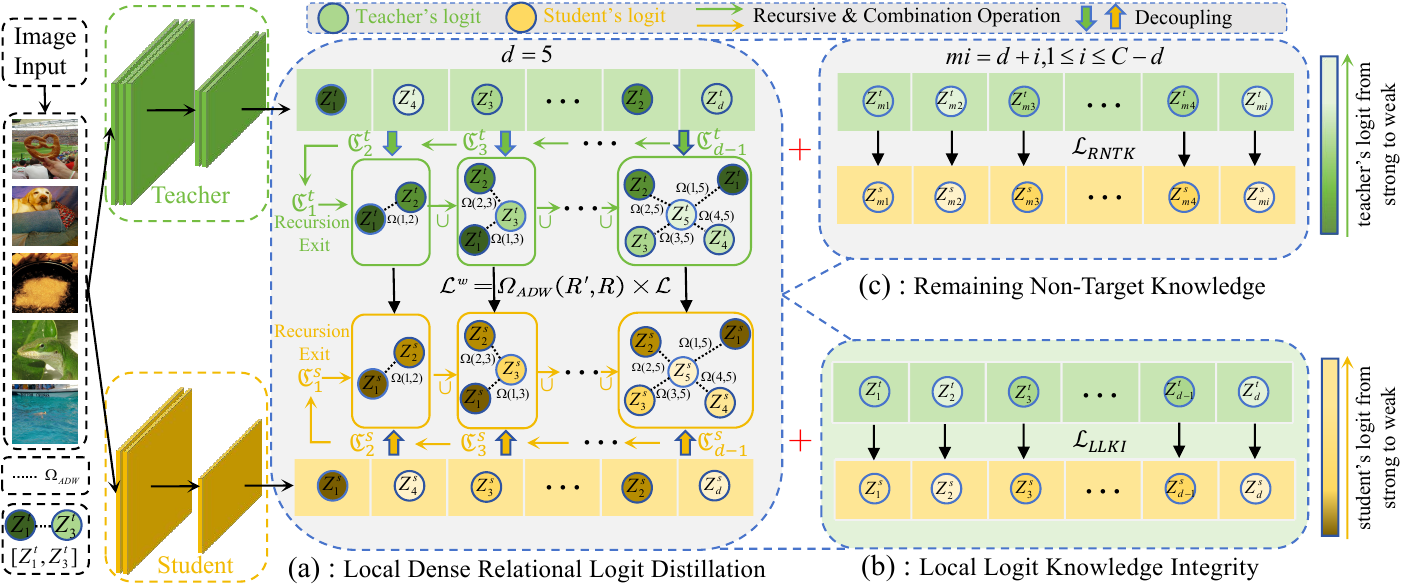}} \vspace{-7pt}
 \caption{
 \label{fig_tt}
      Overview of the proposed LDRLD framework, which includes the three key loss functions:  $\mathcal{L}^{w}$, $\mathcal{L}_{RNTK}$, and $\mathcal{L}_{LLKI}$} 
 \end{figure*}

\noindent\textbf{Notations.}~For convenience, we consider the $n$-th training sample $x_n$ with label $y_n$.~Let $C$ denote the number of categories. The ordered logit value vectors for the teacher and the student are defined as $\mathbf{Z}^t=\left[\mathbf{Z}_1^t, \mathbf{Z}_2^t, \ldots, \mathbf{Z}_C^t\right] \in \mathbb{R}^{1\times C}$ and $\mathbf{Z}^s=\left[\mathbf{Z}_1^s, \mathbf{Z}_2^s, \ldots, \mathbf{Z}_C^s\right] \in \mathbb{R}^{1\times C}$, respectively.  Here, $\mathbf{Z}_j^t$ and $\mathbf{Z}_j^s$, where $j \neq k$, represent the logits for the non-target classes,and $k$ is the index of $y_n$, as discussed in~\cite{zhao2022decoupled}.

\subsection{Vanilla Knowledge Distillation}
\label{ckd}
In KD~\cite{hinton2015distilling}, knowledge is transferred by minimizing the Kullback-Leibler ($\mathcal{KL}$) divergence $\mathcal{L}_{KD}$ between the student's and teacher's output probability distributions.~The output probability distributions for the teacher and the student are represented as $P^t=\left[p_1^t, p_2^t, \ldots, p_C^t\right] \in \mathbb{R}^{1\times C}$ and $P^s=\left[p_1^s, p_2^s, \ldots, p_C^s\right]\in \mathbb{R}^{1\times C}$, respectively. Here, $p_j^t$ and $p_j^s$ represent the predicted probabilities for the $j$-th class of the teacher and the student, respectively, computed as:
\begin{equation}
\begin{aligned}
p_j^t=\exp \left(\mathbf{Z}_j^t\right)/S^{t},\quad p_j^s=\exp \left(\mathbf{Z}_j^s\right)/S^{s},
\end{aligned}
\end{equation}
where $S^{t}$ and $S^{s}$ are normalization factors computed as $S^{t} = \sum_{i=1}^C \exp(\mathbf{Z}_i^t)$ and $S^{s} = \sum_{i=1}^C \exp(\mathbf{Z}_i^s)$, respectively, with temperature omitted for simplicity.
Additionally, cross-entropy is taken as
incorporate the task loss $\mathcal{L}_{Task}$ to ensure accurate predictions. The total loss is thus formulated as:
\begin{equation}
\begin{aligned}
\mathcal{L}_{Total}
&=\mathcal{L}_{Task}+\gamma\mathcal{L}_{KD}\\
&=- \sum_{j=1}^C y_j \log p_j^s +\gamma \sum_{j=1}^C p_j^t \log ( p_j^t/p_j^s),
\end{aligned}
\end{equation}
where $\gamma $ serves as the weight to balance the two losses. 
\subsection{ Local Dense Relational Logit Distillation}
\label{LRD}

To provide richer information for student learning, we propose the LDRLD method shown in Fig.~\ref{fig_tt}(a).~Our method recursively decouples and recombines logit knowledge, capturing fine-grained logit relationships with greater inter-class distinction.~Specifically, we define the recursion depth $d$ by selecting the top-$d$ logits according to student's logit arrange for a given sample, with $d \le C$.~The selected logits for the teacher and the student are denoted as $\mathcal{H}^t=[\mathbf{Z}_1^t, \mathbf{Z}_2^t, \ldots, \mathbf{Z}_d^t] \subseteq \mathbf{Z}^t$ and $\mathcal{H}^s=[\mathbf{Z}_1^s, \mathbf{Z}_2^s, \ldots, \mathbf{Z}_d^s]\subseteq \mathbf{Z}^s$, respectively, with each sequence arranged in descending order.~Additionally, the argmax is used to identify the index of the maximum logit from the student's output, $\mathbf{Z}^s$.~To ensure consistent alignment, a mask is applied to both the student's and teacher's outputs.~The mask vector $\mathbf{M} \in\mathbb{R}^C$ in the $d$-th iteration is defined as:
\vspace{-2pt}
\begin{equation}
\begin{aligned}
\mathbf{M}_{\pi(d)}= \begin{cases}-\infty & \text { if } \pi(d)=i_{\max}, \\ 1 & \text { otherwise, }\end{cases}
\end{aligned}
\vspace{-4pt}
\label{ma1}
\end{equation}
where $
i_{\max }=\arg \max _i \mathbf{Z}_i^s~(i=1,...,d)$ and $\pi(d)$ denotes the index of the $d$-th largest student's logit output.
The proposed LDRLD method works as follows:

\textbf{Step 1: Extraction of the Maximum Logit and the Masking Operation.} The maximum logit is extracted from the vectors $\mathbf{Z}^t$ and $\mathbf{Z}^s$ as follows:
\vspace{-2pt}
\begin{equation}
\begin{aligned}
 \mathbf{Z}_1^t=  \mathbf{Z}_{\pi(1)}^t,\quad \mathbf{Z}_1^s= \mathbf{Z}_{\pi(1)}^s.
\end{aligned}
\label{mask7}
\vspace{-3pt}
\end{equation}

Based on Eq.~(\ref{ma1}), the mask $\mathbf{M}_{\pi(1)}$ is then applied to update the teacher's and student's logit vectors as follows:
\vspace{-6pt}
\begin{equation}
\begin{aligned}
\mathbf{Z}^t \gets \mathbf{M}_{\pi(1)}\odot\mathbf{Z}^t, \quad \mathbf{Z}^s \gets \mathbf{M}_{\pi(1)}\odot\mathbf{Z}^s,
\end{aligned}
\label{mask1}
\vspace{-6pt}
\end{equation}
where $\odot$ denotes the Hadamard product.~The resulting logits are used in \textbf{Step 2}.

\textbf{Step 2: Recursive Decoupling and Combination Operation.}~We define the concatenation function $\phi : \mathbb{R} \uplus \mathbb{R} \rightarrow \mathbb{R}^{1\times2}$, where $\uplus$ denotes an operation that combines two logits into a pair.~Similar to \textbf{Step 1}, we use Eqs.~(\ref{ma1}) and (\ref{mask7}) to extract the second-largest student's logit when $d=2$, denoted as $\mathbf{Z}_2^t$ and $\mathbf{Z}_2^s$, respectively.~The mask $\mathbf{M}_{\pi(2)}$ based on Eq.~(\ref{ma1}) is then applied to update these logit vectors:
 \vspace{-5pt}
\begin{equation}
\begin{aligned}
 \mathbf{Z}^t \gets \mathbf{M}_{\pi(2)}\odot\mathbf{Z}^t, \quad \mathbf{Z}^s \gets \mathbf{M}_{\pi(2)}\odot\mathbf{Z}^s.
\end{aligned}
\vspace{-5pt}
\end{equation}

After that, ~the combination generation proceeds recursively as described below:

\textbf{Base Case.} For depth $d=2$, the first set of combinations is generated for the teacher and the student:
\vspace{-5pt}
\begin{equation}
\begin{aligned}
\mathfrak{C}_{1}^{t} = \phi(\mathbf{Z}_1^t, \mathbf{Z}_2^t), \quad \mathfrak{C}_{1}^{s} = \phi(\mathbf{Z}_1^s, \mathbf{Z}_2^s),
\end{aligned}
\vspace{-5pt}
\end{equation}
where the symbol $\mathfrak{C}$ represents the result of the combination, and $\mathfrak{C}_0=\emptyset$. $\mathfrak{C}_1^t$ and $\mathfrak{C}_1^s$ denote the recursive exit.

\textbf{Recursive Step.}  We repeat \textbf{Steps 1 and 2} as follows:
\begin{itemize}
\item (1) Extraction: Select the \(d\)-th largest logit:
\vspace{-5pt}
\begin{equation}
\begin{aligned}
 \mathbf{Z}_d^t=  \mathbf{Z}_{\pi(d)}^t,\quad \mathbf{Z}_d^s= \mathbf{Z}_{\pi(d)}^s,
\end{aligned}
\label{mask8}
\vspace{-5pt}
\end{equation}
where $\pi(d)$ is the index of the $d$-th largest student's logit output, $\mathbf{Z}^s$. 
\item (2) Masking:~Apply the mask $\mathbf{M}_{\pi(d)}$ based on Eq.~(\ref{ma1}) to update
the teacher’s and student’s logit vectors:
\vspace{-6pt}
\begin{equation}
\begin{aligned}
\mathbf{Z}^t \gets \mathbf{M}_{\pi(d)} \odot \mathbf{Z}^t, \quad \mathbf{Z}^s \gets \mathbf{M}_{\pi(d)} \odot \mathbf{Z}^s.
\end{aligned}
\vspace{-6pt}
\end{equation}
\item (3) Combination Generation: 
For any recursion depth $d \ge 2$, new sets of combinations are generated by applying $\phi$ to the newly extracted logit $\mathbf{Z}_d^t$ and $\mathbf{Z}_d^s$, and all previously extracted logits ($\forall i, 2 \leq i \leq d-1$). The combination sets $\mathfrak{C}_{d-1}^t$ and $\mathfrak{C}_{d-1}^s$ are defined as follows:
\end{itemize}
\vspace{-1.0pt}
\begin{equation} 
\mathfrak{C}_{d-1}^t=\left\{\mathfrak{C}_{d-2}^t \cup \bigcup_{i=1}^{d-1}\left\{\phi(\mathbf{Z}_i^t, \mathbf{Z}_d^t)\right\}\right\}\in \mathbb{R}^{{\frac{d(d-1)}{2}\times2}},
\end{equation}
\begin{equation}
\mathfrak{C}_{d-1}^s=\left\{\mathfrak{C}_{d-2}^s \cup \bigcup_{i=1}^{d-1}\left\{\phi\left(\mathbf{Z}_i^s, \mathbf{Z}_d^s\right)\right\}\right\}\in \mathbb{R}^{{\frac{d(d-1)}{2}\times2}},
\end{equation}
where $\bigcup$ represents the union operation.

\textbf{Step 3: Normalization.}~To ensure that the combinations in $\mathfrak{C}$ satisfy the properties of a probability distribution, we normalize them using the softmax function $\sigma$.~Let $(p_i^t, p_j^t)$ denote the teacher's logit pair, and let $(p_i^s, p_j^s)$ denote the student's logit pair. The probabilities $p_i^t, p_j^t, p_i^s$, and $p_j^s$ are computed as follows via  $\sigma$:
\begin{equation}
    \left\{ \begin{array}{l}
	p_{i}^{t}=\exp \left( \mathbf{Z}_{i}^{t}/\tau \right) /S_{ij}^{t}, p_{j}^{t}=\exp \left( \mathbf{Z}_{j}^{t}/\tau \right) /S_{ij}^{t},\vspace{5pt}\\
	p_{i}^{s}=\exp \left( \mathbf{Z}_{i}^{s}/\tau \right) /S_{ij}^{s}, p_{j}^{s}=\exp \left( \mathbf{Z}_{j}^{s}/\tau \right) /S_{ij}^{s},\\
\end{array} \right. 
\end{equation}
where $S^{t}_{ij}$ and $S^{s}_{ij}$ are the normalization factors for the teacher and student, respectively.~Here, $S^{t}_{ij} =\exp \left(\mathbf{Z}_i^t/\tau\right)+\exp \left(\mathbf{Z}_j^t/\tau\right)$ and $S^{s}_{ij} =\exp \left(\mathbf{Z}_i^s/\tau\right)+\exp \left(\mathbf{Z}_j^s/\tau\right)$. $\tau$ represents the temperature coefficient that controls the softness of the output probability distribution.

Building upon \textbf{Steps 1 to 3}, we transfer the teacher's \textit{local dense relational logit knowledge} to the student using the loss defined as follows:
\vspace{-4pt}
\begin{equation}
\adjustbox{scale=0.9}{$
\begin{aligned}
\mathcal{L} 
&= \sum_{i=1}^{d-1} \mathcal{KL}\left(\mathfrak{C}_{i}^t, \mathfrak{C}_{i}^s\right) \\
&= \sum_{i=1}^{d-1} \sum_{j=i+1}^{d} \mathcal{KL}\left(\sigma(\phi(\mathbf{Z}_i^t, \mathbf{Z}_j^t),\tau), \sigma(\phi(\mathbf{Z}_i^s, \mathbf{Z}_j^s),\tau)\right)\\
&= \sum_{i=1}^{d-1} \sum_{j=i+1}^{d} \left[ p_i^t \log (p_i^t/p_i^s) + p_j^t \log (p_j^t/p_j^s) \right]. 
\end{aligned}
$}
\label{rkdd}
\vspace{-4pt}
\end{equation}
\textbf{Local Logit Knowledge Integrity.} Additionally, when the recursion depth 
$d$ is reached, the $d$ logits are generatd sequentially to capture hierarchical relationships, thereby preserving local logit knowledge integrity (LLKI).~The resulting teacher's and student's logits, $\mathcal{H}^t$ and $\mathcal{H}^s$, are presented in Fig.~\ref{fig_tt}(b).~The loss $\mathcal{L}_{LLKI}$ is defined as follows:
\begin{equation} 
\begin{aligned}
\mathcal{L}_{LLKI}=\mathcal{KL}(\mathcal{H}^t,\mathcal{H}^s)=\sum_{i=1}^d p_i^t \log \left(p_i^t/p_i^s\right),
\end{aligned}
\end{equation} 
where the probabilities of the teacher and the student are denoted as $
p_i^t=\sigma(\mathbf{Z}_i^t/\tau)$, and  $p_i^s=\sigma(\mathbf{Z}_i^s/\tau)$, respectively.

\noindent\textbf{Adaptive Decay Weight strategy.}~So far, we have uniformly assigned weights to all category pairs. However, this may hinder the optimization of visually complex or critical categories and lead to network performance degradation. To address this, we propose an Adaptive Decay Weight (ADW) strategy based on human perception of category similarity.~The ADW integrates Inverse Rank Weighting and Exponential Rank Decay.~It dynamically adjusts weights to enhance discrimination by assigning greater weights to category pairs that are more difficult for the model to distinguish (e.g., $[\mathbf{Z}_1^t, \mathbf{Z}_2^t]$), and smaller weights to less related category pairs (e.g., $[\mathbf{Z}_1^t, \mathbf{Z}_4^t]$).~This targeted weighting sharpens the student's focus on key categories while reducing attention to others.~Details of the two key components of the ADW strategy are described below:

\textbf{(1) Inverse Rank Weighting.}~To emphasize the importance of closely ranked category pairs, we introduce Inverse Rank Weighting (IRW), which is defined as:
\begin{equation} 
\begin{aligned}
\Gamma_{IRW}(R^{\prime},R)=Inv(\left | R-R^{\prime} \right | +\epsilon),
\end{aligned}
\label{IRW}
\end{equation} 
where $Inv(x)$ denotes the inverse of $x$. $R^{\prime}$ and $R$ are the rankings of categories $i$ and $j$ in the pair $[\mathbf{Z}_i, \mathbf{Z}_j]$, respectively. The term $\left|R-R^{\prime}\right|$ represents the difference in category rankings, and $\epsilon=1.50$ prevents division by zero.

IRW assigns larger weights to pairs with smaller ranking differences and smaller weights to those with larger differences. Thus, IRW directs the student's focus toward optimizing more similar category pairs.

\textbf{(2) Exponential Rank Decay.} When calculating the pairwise weights according to Eq.~(\ref{IRW}), note that the pairs $[\mathbf{Z}_1^t, \mathbf{Z}_2^t]$ and $[\mathbf{Z}_{12}^t, \mathbf{Z}_{13}^t]$  receive the same weight. However, intuitively, distinguishing between $[\mathbf{Z}_1^t, \mathbf{Z}_2^t]$ is more difficult due to highly semantic similarity, so the weight for the former should be greater than that for the latter. Based on this understanding, we propose an Exponential Ranking Decay (ERD) that dynamically adjusts weight decay according to the sum of the rankings, defined as:
\begin{equation} 
\begin{aligned}
\Phi_{ERD}(R^{\prime},R)= \delta \times \exp(-\lambda (R+R^{\prime})),
\end{aligned}
\end{equation} 
where $\delta$, the weight, is set to 2.0, and $\lambda$, the decay rate, is set to 0.05 in our experiments.~The decay rate controls how quickly the weight decreases as the sum of rankings $(R^{\prime} + R)$ increase, and a larger value of $\lambda$ results in faster weight decay.

With IRW and ERD, our ADW strategy is defined as follows:
\begin{equation} 
\begin{aligned}
\Omega_{ADW}(R^{\prime}, R)=\Gamma_{IRW}(R^{\prime}, R)\times \Phi_{ERD}(R^{\prime}, R).
\label{ADW}
\end{aligned}
\end{equation} 

By combining Eq.~(\ref{rkdd}) and Eq.~(\ref{ADW}), we formulate the loss of $\mathcal{L}^{w}$ to facilitate the transfer of local dense relational logit knowledge, defined as follows:
\vspace{-4pt}
\begin{equation}
\adjustbox{scale=0.9}{$
\begin{aligned}
\mathcal{L}^{w}=\sum_{i=1}^{d-1}{\sum_{j=i+1}^d{\Omega _{ADW}}}\left( i,j \right) \left[ p_i^t \log (p_i^t/p_i^s) + p_j^t \log (p_j^t/p_j^s) \right].
\end{aligned} 
$}
\end{equation} 

The complete local logit knowledge $\mathcal{L}_{Local}$ is defined as follows:
\begin{equation}
\begin{aligned}
\mathcal{L}_{Local}=\mathcal{L}^{w}+\mathcal{L}_{LLKI}.
\end{aligned} 
\label{local}
\end{equation} 
\textbf{Remaining  Non-Target Knowledge.}
Inspired by the effectiveness of non-target knowledge in previous studies~\cite{ding2021knowledge,zhao2022decoupled,yang2023knowledge,li2022asymmetric}, we leverage the remaining non-target classes to ensure comprehensive knowledge transfer and improve student performance.~We denote the remaining logits for the teacher and student as $\bar{\mathcal{H}}^t=[\mathbf{Z}_{d+1}^t,\mathbf{Z}_{d+2}^t,\ldots,\mathbf{Z}_{C}^t]$, and $\bar{\mathcal{H}}^s=[\mathbf{Z}_{d+1}^s,\mathbf{Z}_{d+2}^s,\ldots,\mathbf{Z}_{C}^s]$, respectively, in Fig.~\ref{fig_tt}(c).~The knowledge is distilled using $\mathcal{L}_{RNTK}$, defined as follows:
\begin{equation} 
\begin{aligned}
\mathcal{L}_{RNTK}=\mathcal{KL}(\bar{\mathcal{H}^t},\bar{\mathcal{H}^s})=\sum_{i=d+1}^C \bar{p}_i^t \log \left(\bar{p}_i^t/\bar{p}_i^s\right),
\end{aligned}
\label{RNTK}
\end{equation}
where the probabilities of the teacher and the student are $\bar{p}_i^t=\sigma(\mathbf{Z}_i^t/\tau)$ and  $\bar{p}_i^s=\sigma(\mathbf{Z}_i^s/\tau)$, where $d+1\leq i\leq C$, respectively.

\subsection{Optimization Objective}
The proposed method integrates task loss, local relational information, and knowledge from  non-target classes. This integration enables precise adjustment of the student's parameters through the following optimization function:
\begin{equation}
\begin{aligned}
 \mathcal{L}_{LDRLD}= \mathcal{L}_{Task} + \alpha \mathcal{L}_{Local} + \beta \mathcal{L}_{RNTK}
\end{aligned}
\label{total_loss}
\end{equation}
where $\alpha$ and $\beta$ are weighting factors that balance the contributions of these losses.

To facilitate the understanding and implementation of our method, we provide a detailed algorithm and specific descriptions in the \textit{Supplementary Material}.

\begin{table*}[!t]
\vspace{-0.2in}
\begin{center}
\setlength\tabcolsep{10pt}
\scalebox{0.75}{
\begin{tabular}{c|c|c|ccccccc}
\tophline
\multirow{2}*{Distillation} &Teacher &\multirow{2}*{Publication} &ResNet56 & ResNet110 & ResNet110 & ResNet32×4 & WRN-40-2 & WRN-40-2 & VGG13 \\
  &Accuracy & & 72.34 & 74.31 & 74.31 & 79.42 & 75.61 & 75.61 & 74.64 \\
\cline{2-10}
\multirow{2}*{Manner} & Student &\multirow{2}*{Year}& ResNet20 & ResNet20 & ResNet32 & ResNet8×4 & WRN-16-2 & WRN-40-1 & VGG8 \\
~ &Accuracy & & 69.06 & 69.06 & 71.14 & 72.50 & 73.26 & 71.98 & 70.36 \\
\hline
 ~& FitNet~\cite{romero2014fitnets} &ICLR'15& 69.21 & 68.99 & 71.06 & 73.50 & 73.58 & 72.24 & 71.02 \\  
& RKD~\cite{park2019relational} &CVPR'19& 69.61 & 69.25 & 71.82 & 71.90 & 73.35 & 72.22 & 71.48 \\ 
~ & CRD~\cite{tian2019contrastive} &ICLR'20& 71.16 & 71.46 & 73.48 & 75.51 & 75.48 & 74.14 & 73.94 \\ 
Features & ReviewKD~\cite{chen2021distilling} &CVPR'21& \textbf{71.89} & \textbf{71.85} & \textbf{73.89} & 75.63 & 76.12 & 75.09 & 74.84 \\
~ & CAT-KD~\cite{guo2023class} &CVPR'23&71.62 &71.14&73.62& \textbf{76.91}& 75.60 &74.82& 74.65 \\
~&FCFD~\cite{liu2023functionconsistent}&ICLR'23&71.68&-&-& 76.80&\textbf{76.34}& \textbf{75.43}&\textbf{74.86}\\
~ & NORM~\cite{liu2023norm} &ICLR'23& 71.35 & 71.55 & 73.67 & 76.49 & 75.65 & 74.82 & 73.95 \\
\hline
& KD~\cite{hinton2015distilling} &NeurIPS'14& 70.66 & 70.67 & 73.08 & 73.33 & 74.92 & 73.54 & 72.98 \\ 
~ & DKD~\cite{zhao2022decoupled} &CVPR'22& 71.97 & 70.91 & 74.11 & 76.32 & 76.24 & 74.81 & 74.68 \\ 
 & CTKD~\cite{li2023curriculum}&AAAI'23 & 71.19 & 70.99 & 73.52 & 73.70 & 75.45 & 73.93 & 73.52 \\
&NKD~\cite{yang2023knowledge}&ICCV'23&71.18&71.26& 73.50&76.35& 75.43&74.25& 74.86\\
&LCKA~\cite{Zhou2024RethinkingCK}&IJCAI'24&-&70.87 & 73.64&75.12&75.78 & 74.63& 74.35\\
Logits&LSKD~\cite{Sun2024Logit}&CVPR'24& 71.24  & 71.61  & 73.76 & 76.16 & 76.22& 74.43 & 74.23\\
&SDD~\cite{luo2024scale}&CVPR'24& 71.52 & 71.58 & 73.97 & 75.09 & 75.86 & 74.53 & 73.91\\
&WTTM~\cite{zheng2024knowledge}&ICLR'24&71.92&71.67&74.13& 76.06&\textbf{76.37}&74.58&74.44\\
&TeKAP~\cite{hossain2025single}&ICLR'25&71.32&71.24*&73.42*& 74.79&75.21&73.80&74.00\\
&LDRLD &Ours&\textbf{72.20}&\textbf{71.98}&\textbf{74.16}& \textbf{77.20}&76.35&\textbf{74.98}&\textbf{75.06}\\
~ & $\Delta$ &Ours& \textbf{\textcolor{deepgreen}{+1.54}} & \textbf{\textcolor{deepgreen}{+1.31}} & \textbf{\textcolor{deepgreen}{+1.08}} & \textbf{\textcolor{deepgreen}{+3.87}} & 
\textbf{\textcolor{deepgreen}{+1.43}} & \textbf{\textcolor{deepgreen}{+1.44}} & \textbf{\textcolor{deepgreen}{+2.08}} \\
\tophline
\end{tabular}}
\vspace{-2pt}
\caption{Evaluation of the top-1 accuracy (\%) of student with the same architecture on the CIFAR-100 validation set.~$\Delta$ denotes the improvement over the vanilla KD method. \textbf{We report the average results of three runs for all tables.} * represents our reproduced results.}
\label{cifar-100-same}
\end{center}
\end{table*}
\section{Experiments}
\begin{table*}[!ht]
\begin{center}
\centering
\setlength\tabcolsep{11.3pt}
\scalebox{0.73}{
\begin{tabular}{c|c|c|cccccc}
\tophline
\multirow{2}*{Distillation}&Teacher  &\multirow{2}*{Publication}& ResNet32$\times$4   & WRN40-2      & ResNet50     & VGG13        & ResNet32$\times$4   & ResNet50      \\ 
&Accuracy      && 79.42        & 75.61        & 79.34        & 74.64        & 79.42        & 79.34         \\ 
\cline{2-9}
\multirow{2}*{Manner}&Student  &\multirow{2}*{Year}& ShufleNetV1  & ShufleNetV1  & MobileNetV2  & MobileNetV2  & ShufleNetV2  & VGG8          \\ 
&Accuracy   && 70.50        & 70.50        & 64.6         & 64.6         & 71.82        & 70.36         \\ 
\hline
&FitNet~\cite{romero2014fitnets}   &ICLR'15& 73.54        & 73.73        & 63.16        & 63.16        & 73.54        & 70.69         \\ 
&RKD~\cite{park2019relational}   &CVPR'19   & 72.28        & 72.21        & 64.43        & 64.43        & 73.21        & 71.50         \\ 
&CRD~\cite{tian2019contrastive}   &ICLR'20   & 75.11        & 76.05        & 69.11        & 69.11        & 75.65        & 74.30         \\
Features&ReviewKD~\cite{chen2021distilling} &CVPR'21& 77.45        & 77.14        & 69.89        & 70.37        & 77.78        & 75.34         \\ 
&NORM~\cite{liu2023norm}  &ICLR'23    &  77.42        & 77.06        & 70.56       & 68.94        & 78.07        & 75.17         \\ 
&FCFD~\cite{liu2023functionconsistent}  &ICLR'23    &  78.12        & \textbf{77.81}        & 71.07       & \textbf{70.67}        & 78.20        &-         \\ 
&CAT-KD~\cite{guo2023class}  &CVPR'23    &   \textbf{78.26}        & 77.35        & \textbf{71.36}       & 69.13        & \textbf{78.41}        & \textbf{75.39}         \\ 
\hline
&KD~\cite{hinton2015distilling}    &NeurIPS'14   & 74.07        & 74.83        & 67.35        & 67.37        & 74.45        & 73.81         \\ 
&DKD~\cite{zhao2022decoupled}    &CVPR'22  & 76.45        & 76.70        & 70.35        & 69.71        & 77.07        & 75.34         \\ 
&CTKD~\cite{li2023curriculum}    &AAAI'23   & 74.48        & 75.78        & 68.47      & 68.46        & 75.31        & 73.63         \\ 
&NKD~\cite{yang2023knowledge} &ICCV'23& 76.31 & 76.46 &    70.22  &   \textbf{70.67} & 76.92 &74.01\\ 
Logits&LSKD~\cite{Sun2024Logit} &CVPR'24& - & - &    69.02  &   68.61 & 75.56 &74.42\\ 
&SDD~\cite{luo2024scale} &CVPR'24& 76.30 & 76.65 &    69.55  &  68.79 & 76.67 &74.89\\ 
&WTTM~\cite{zheng2024knowledge} &ICLR'24& 74.37 & 75.42 &   69.59  &  69.16 &76.55 &74.82\\ 
&TeKAP~\cite{hossain2025single} &ICLR'25& 74.92 & 76.75 &  69.00*  &  67.39* &75.43 &74.37*\\
&LDRLD &Ours&\textbf{76.46}& \textbf{77.09} &  \textbf{70.74}&70.11& \textbf{77.33}& \textbf{75.49} \\ 
&$\Delta$ &Ours& \textbf{\textcolor{deepgreen}{+2.39}}&  \textbf{\textcolor{deepgreen}{+2.26}} &  \textbf{\textcolor{deepgreen}{+3.39}}& \textbf{\textcolor{deepgreen}{+2.74}}&  \textbf{\textcolor{deepgreen}{+2.88}}&  \textbf{\textcolor{deepgreen}{+1.68}} \\  
\bottomline
\end{tabular}
}
\end{center}
\vspace{-14pt}
\caption{Evaluation of the top-1 accuracy (\%) of student with the different architecture on the CIFAR-100 validation set.}
\label{cifar-100-diff}
\end{table*}

\begin{table*}[!ht]
  \centering
\setlength\tabcolsep{0.3pt}
\renewcommand{\arraystretch}{1.80}
\scalebox{0.77}{
    \begin{tabular}{|c|c|c|c|c|c|c|c|c|c|c|c|c|c|c|c|c|}
    \tophline
    \multicolumn{16}{|c|}{ResNet34 (teacher): 73.31\% Top-1, 91.42\% Top-5 accuracy.
ResNet18 (student): 69.75\% Top-1, 89.07\% Top-5 accuracy.} \\
    \tophline
    \textbf{Features} & \multicolumn{1}{c|}{AT\cite{komodakis2017paying}} & \multicolumn{1}{c|}{CRD\cite{tian2019contrastive}} & \multicolumn{1}{c|}{ReviewKD\cite{chen2021distilling}} & \multicolumn{1}{c|}{FCFD\cite{liu2023functionconsistent}} & \multicolumn{1}{c|}{CAT-KD\cite{guo2023class}} & \textbf{Logits} &      \multicolumn{1}{c|}{KD\cite{hinton2015distilling}} & \multicolumn{1}{c|}{CTKD\cite{li2023curriculum}} & \multicolumn{1}{c|}{DKD\cite{zhao2022decoupled}}  & \multicolumn{1}{c|}{LSKD\cite{Sun2024Logit}} & \multicolumn{1}{c|}{SDD\cite{luo2024scale}}& 
    \multicolumn{1}{c|}{WTTM\cite{zheng2024knowledge}}&\multicolumn{1}{c|}{TeKAP\cite{hossain2025single}}& \multicolumn{1}{c|}{LDRLD}&\multicolumn{1}{c|}{$\Delta$}\\
    \midhline
    Top-1  & \multicolumn{1}{c|}{70.69}& \multicolumn{1}{c|}{71.17} & \multicolumn{1}{c|}{71.61}  & \multicolumn{1}{c|}{\textbf{72.24}} & \multicolumn{1}{c|}{71.26}& Top-1 & \multicolumn{1}{c|}{70.66} & \multicolumn{1}{c|}{71.32} & \multicolumn{1}{c|}{71.70} & \multicolumn{1}{c|}{71.42} & \multicolumn{1}
    {c|}{71.44} & \multicolumn{1}
    {c|}{\textbf{72.19}}& \multicolumn{1}
    {c|}{71.35*}&\multicolumn{1}
    {c|}{71.88}&\multicolumn{1}
    {c|}{\textbf{\textcolor{deepgreen}{+1.22}}}\\
    \midhline
    Top-5  & \multicolumn{1}{c|}{90.01} & \multicolumn{1}{c|}{90.13} & \multicolumn{1}{c|}{90.51} & \multicolumn{1}{c|}{\textbf{90.74}}  & \multicolumn{1}{c|}{90.45}& Top-5 & \multicolumn{1}{c|}{89.88} & \multicolumn{1}{c|}{90.27} & \multicolumn{1}{c|}{90.31} & \multicolumn{1}{c|}{90.29} & \multicolumn{1}
    {c|}{90.05} & \multicolumn{1}{c|}{-} & \multicolumn{1}{c|}{90.54*}&\multicolumn{1}{c|}{\textbf{90.58}} &\multicolumn{1}
    {c|}{\textbf{\textcolor{deepgreen}{+0.70}}}\\
    \bottomline
    \multicolumn{16}{|c|}{ResNet50 (teacher): 76.16\% Top-1, 92.87\% Top-5 accuracy.
MobileNetV1 (student): 68.87\% Top-1, 88.76\% Top-5 accuracy.} \\
    \tophline
    \textbf{Features} & \multicolumn{1}{c|}{AT\cite{komodakis2017paying}}  & \multicolumn{1}{c|}{CRD\cite{tian2019contrastive}} & \multicolumn{1}{c|}{ReviewKD\cite{chen2021distilling}} & \multicolumn{1}{c|}{FCFD\cite{liu2023functionconsistent}}  & \multicolumn{1}{c|}{CAT-KD\cite{guo2023class}} & \textbf{Logits} &      \multicolumn{1}{c|}{KD\cite{hinton2015distilling}} & \multicolumn{1}{c|}{IPWD\cite{niu2022respecting}} & \multicolumn{1}{c|}{DKD\cite{zhao2022decoupled}}  & \multicolumn{1}{c|}{LSKD\cite{Sun2024Logit}} & \multicolumn{1}{c|}{SDD\cite{luo2024scale}}& 
    \multicolumn{1}{c|}{WTTM\cite{zheng2024knowledge}}& 
    \multicolumn{1}{c|}{TeKAP\cite{hossain2025single}}&
    \multicolumn{1}{c|}{LDRLD}&\multicolumn{1}{c|}{$\Delta$}\\
    \midhline
    Top-1  & \multicolumn{1}{c|}{70.18}  & \multicolumn{1}{c|}{71.32} & \multicolumn{1}{c|}{72.56}  & \multicolumn{1}{c|}{\textbf{73.37}} & \multicolumn{1}{c|}{72.24}& Top-1 & \multicolumn{1}{c|}{70.49} & \multicolumn{1}{c|}{72.65} & \multicolumn{1}{c|}{72.05} &  \multicolumn{1}{c|}{72.18} & \multicolumn{1}{c|}
    {72.24} & \multicolumn{1}{c|}{73.09}& \multicolumn{1}{c|}{72.87*}&\multicolumn{1}{c|}{\textbf{73.12}}&\multicolumn{1}{c|}{\textbf{\textcolor{deepgreen}{+2.63}}}\\
    \midhline
    Top-5  & \multicolumn{1}{c|}{89.68} & \multicolumn{1}{c|}{90.41} & \multicolumn{1}{c|}{91.00}  & \multicolumn{1}{c|}{\textbf{91.35}}  & \multicolumn{1}{c|}{91.13}& Top-5 & \multicolumn{1}{c|}{89.92} & \multicolumn{1}{c|}{91.08} & \multicolumn{1}{c|}{91.05} & \multicolumn{1}{c|}{90.80} & \multicolumn{1}{c|}
    {90.71} & \multicolumn{1}{c|}{-}& \multicolumn{1}{c|}{91.05*} & \multicolumn{1}{c|}{\textbf{91.43}}&\multicolumn{1}{c|}{\textbf{\textcolor{deepgreen}{+1.51}}}\\
    \bottomline
    \end{tabular}%
    }
    \vspace{-0.2cm}
     \caption{Evaluate the top-1 and top-5 accuracy (\%) of student using same and different architectures on the ImageNet-1K validation set.}
  \label{tab:imagenet}%
  \vspace{-6pt}
\end{table*}

\subsection{Comparison with State-of-the-Art Methods}
\vspace{-2pt}
\textbf{Image Classification Results on CIFAR-100.} We evaluated our method across various network architectures.~As shown in Table \ref{cifar-100-same}, we first employed an experimental setup where the teacher and the student share the same type of architecture.~In contrast, in Table \ref{cifar-100-diff}, the teacher and the student use different-architectures.~Experimental results show that our proposed method outperforms existing state-of-the-art logit-based KD methods, such as SDD, WTTM, and TeKAP. Specifically, in the same-architecture experiment, our method improves performance by 1.08\% to 3.87\%.~In the different-architecture experiment, the improvement ranges from 1.68\% to 3.39\%.~These improvements are mainly due to LDRLD's ability to extract fine-grained inter-class knowledge from the teacher.~While current logit-based KD methods utilize global softmax to capture inter-class relationships, they may obscure subtle differences among categories, thus impairing targeted learning enhancements. Furthermore, the global softmax mechanism often neglects low-probability categories, limiting performance.~These differences highlight LDRLD's effectiveness, which leverages local logit relational techniques to overcome these limitations, enhance inter-class discriminability and ensure better recognition.

\textbf{Image Classification Results on ImageNet-1K.} The results on ImageNet-1K are shown in Table~\ref{tab:imagenet}.~Using the same type of architecture, our method outperforms the vanilla KD by 1.22\% in top-1 accuracy.~Additionally, when using ResNet50 as the teacher and MobileNetV1 as the student, experimental results show that our method achieves improvements of 2.63\% in top-1 and 1.51\% in top-5 accuracy.~These results further confirm the effectiveness and potential of LDRLD.
However, 
when comparing LDRLD with WTTM using the same architecture, LDRLD's performance is slightly lower.~This is probably because WTTM trains the student to approximate smoother soft targets rather than peaked soft targets, which helps it absorb more knowledge. In contrast, LDRLD faces challenges caused by the complex category structure in the ImageNet-1K and the limited capacity of the student, which make it less effective in capturing relationships between categories.

\begin{table*}[!ht]
  \centering
  \renewcommand{\arraystretch}{1.10}
  \resizebox{0.99\linewidth}{!}{%
    \begin{tabular}{cc|cc|cccc|ccccccc}
      \tophline
      \multirow{2}{*}{Teacher} & \multirow{2}{*}{Student} & \multicolumn{2}{c|}{From Scratch} & \multicolumn{4}{c|}{Feature-based} & \multicolumn{7}{c}{Logit-based} \\
      \cmidrule(lr){3-15} 
                               &                          & T:Accuracy    & S:Accuracy    & FitNet~\cite{romero2014fitnets} & RKD~\cite{park2019relational}   & CRD~\cite{tian2019contrastive}   & 
                               FOFA~\cite{lin2025feature}   & KD~\cite{hinton2015distilling}    & DKD~\cite{zhao2022decoupled}   & DIST~\cite{huang2022knowledge}  & OFA~\cite{hao2023one}  &TeKAP~\cite{hossain2025single}&   LDRLD &$\Delta$ \\
      \midhline
      Swin-T     & ResNet18     & 89.26 & 74.01 & 78.87 & 74.11 & 77.63&81.22 & 78.74 & 80.26 & 77.75 & 80.54&81.38*&\textbf{82.17} &{\textbf{\textcolor{deepgreen}{+3.43}}}\\
      ViT-S      & ResNet18     & 92.04 & 74.01 & 77.71 & 73.72 & 76.60 &80.11 & 77.26& 78.10 & 76.49 &  80.15&79.06*&\textbf{80.36}&{\textbf{\textcolor{deepgreen}{+3.10}}} \\
      Mixer-B/16 & ResNet18     & 87.29 & 74.01 & 77.15 & 73.75 & 76.42 & 80.07&77.79 & 78.67 & 76.36 &  79.39 &80.05*&\textbf{80.69}&{\textbf{\textcolor{deepgreen}{+2.90}}}\\
      Swin-T     & MobileNetV2     & 89.26 & 73.68 & 74.28 & 69.00 & 79.80 & 78.78&74.68 & 71.07 & 72.89 &  80.98 &80.23*&\textbf{81.64}&{\textbf{\textcolor{deepgreen}{+6.96}}}\\
      ViT-S      & MobileNetV2     & 92.04 & 73.68 & 73.54 & 68.46 & 78.14 & 78.87&72.77 & 69.80 & 72.54 &  78.45&78.41* &\textbf{79.21}&{\textbf{\textcolor{deepgreen}{+6.44}}}\\
      Mixer-B/16 & MobileNetV2     & 87.29 & 73.68 & 73.78 & 68.95 & 78.15 &78.62& 73.33 & 70.20 & 73.26 &  78.78& 79.89*&\textbf{80.64}&{\textbf{\textcolor{deepgreen}{+7.31}}}\\
      \midhline
      ConvNeXt-T & DeiT-T  & 88.41 & 68.00 & 60.78 & 69.79 & 65.94 &79.59& 72.99 & 74.60 & 73.55 & 75.76 &76.32*&\textbf{77.46}&{\textbf{\textcolor{deepgreen}{+4.47}}}\\
      Mixer-B/16 & DeiT-T  & 87.29 & 68.00 & 71.05 & 69.89 & 65.35 &74.66& 71.36 & 73.44 & 71.67 &  73.90&74.83*& \textbf{75.31}&{\textbf{\textcolor{deepgreen}{+3.95}}}\\
      ConvNeXt-T & Swin-P  & 88.41 & 72.63 & 24.06 & 71.73 & 67.09&80.74 & 76.44 & 76.80 & 76.41 &   78.32&79.18*& \textbf{80.71}&{\textbf{\textcolor{deepgreen}{+4.27}}}\\
      Mixer-B/16 & Swin-P  & 87.29 & 72.63 & 75.20 & 70.82 & 67.03&78.44 & 75.93 & 76.39 & 75.85 &   78.93&78.97* &\textbf{80.52}&{\textbf{\textcolor{deepgreen}{+4.59}}}\\
      \midhline
      ConvNeXt-T & ResMLP-S12   & 88.41 & 66.56 & 45.47 & 65.82 & 63.35&83.50 & 72.25 & 73.22 & 71.93 &  75.21&\textbf{81.14*}&79.28 &{\textbf{\textcolor{deepgreen}{+7.03}}}\\
      Swin-T     & ResMLP-S12   & 89.26 & 66.56 & 63.12 & 64.66 & 61.72&80.94 & 71.89 & 72.82 & 11.05 & 73.58&80.22*&\textbf{80.54}&{\textbf{\textcolor{deepgreen}{+8.65}}} \\
      \bottomline
    \end{tabular}
  }
    \caption{Evaluation of the top-1 accuracy (\%) of student using ViT-based heterogeneous architectures on the CIFAR100 dataset.}
    \label{tab:cifar}
\end{table*}

\textbf{Image Classification Results on CIFAR-100 using ViT-based Architectures.}~Classical KD methods are typically validated within CNN-based frameworks, but their performance in ViT-based frameworks has been shown to be limited, indicating that their generalization ability remains uncertain. To justify the generalization ability of LDRLD, we conducted ViT-based experiments and employed different architectures for distillation, using students based on CNN, ViT, and MLP, and \textbf{state-of-the-art} results were obtained, and demonstrating the effectiveness of LDRLD, as shown in Table \ref{tab:cifar}. Evaluating LDRLD on ViT-based models provides additional insights into the method's versatility and robustness across different model paradigms.

\subsection{Ablation Studies}
To ensure a fair comparison and validate the effectiveness of component, we conducted ablation experiments on various teacher-student pairs. For more ablation experiments, see the \textit{supplementary materials}.

\begin{figure}[htbp]
    \centering
    \begin{subfigure}{0.235\textwidth}
        \centering
\includegraphics[width=\textwidth]{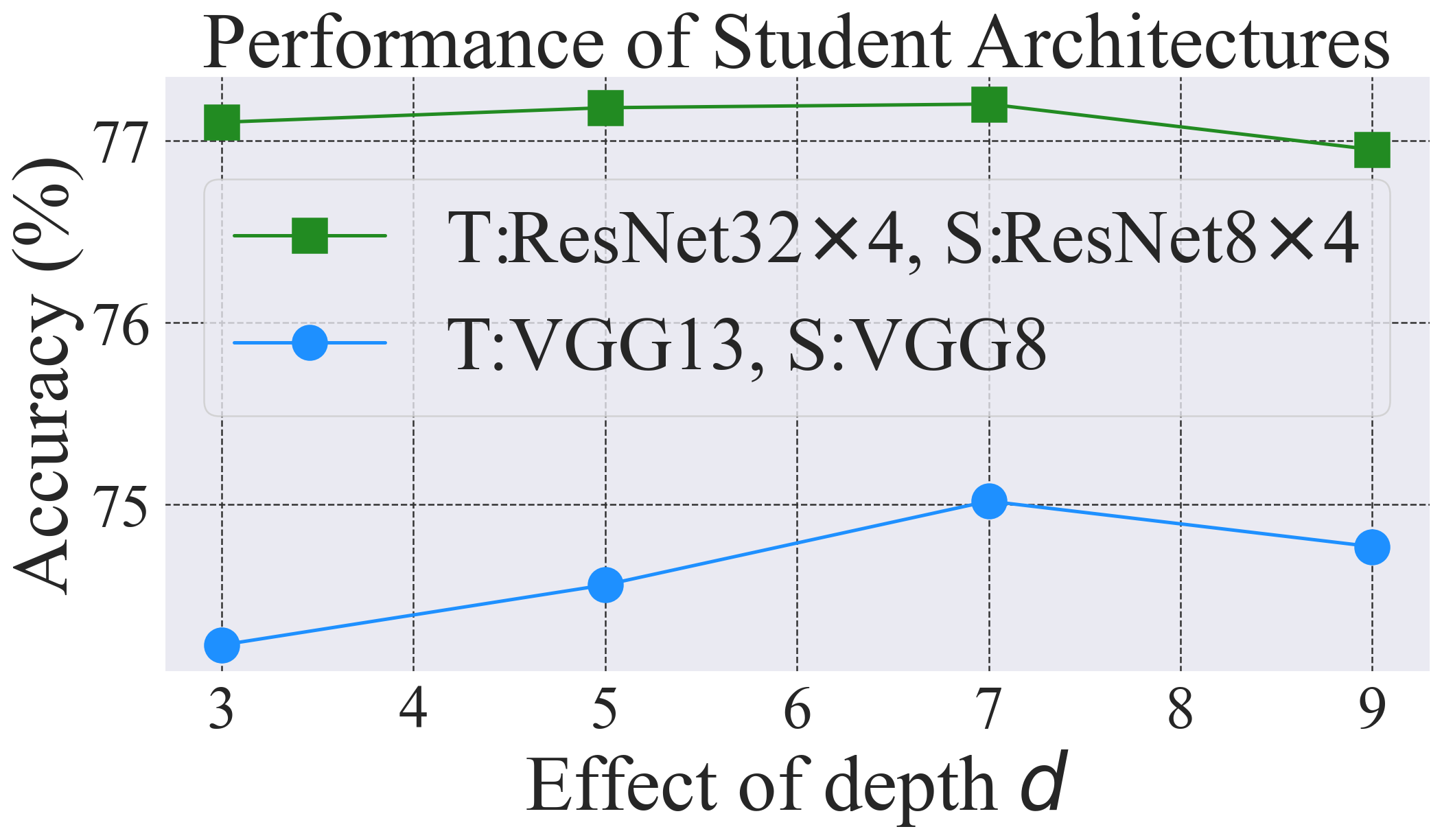}
\vspace{-18pt}
        \caption{Homogeneous architectures}
    \end{subfigure}
    \begin{subfigure}{0.235\textwidth}
        \centering
\includegraphics[width=\textwidth]{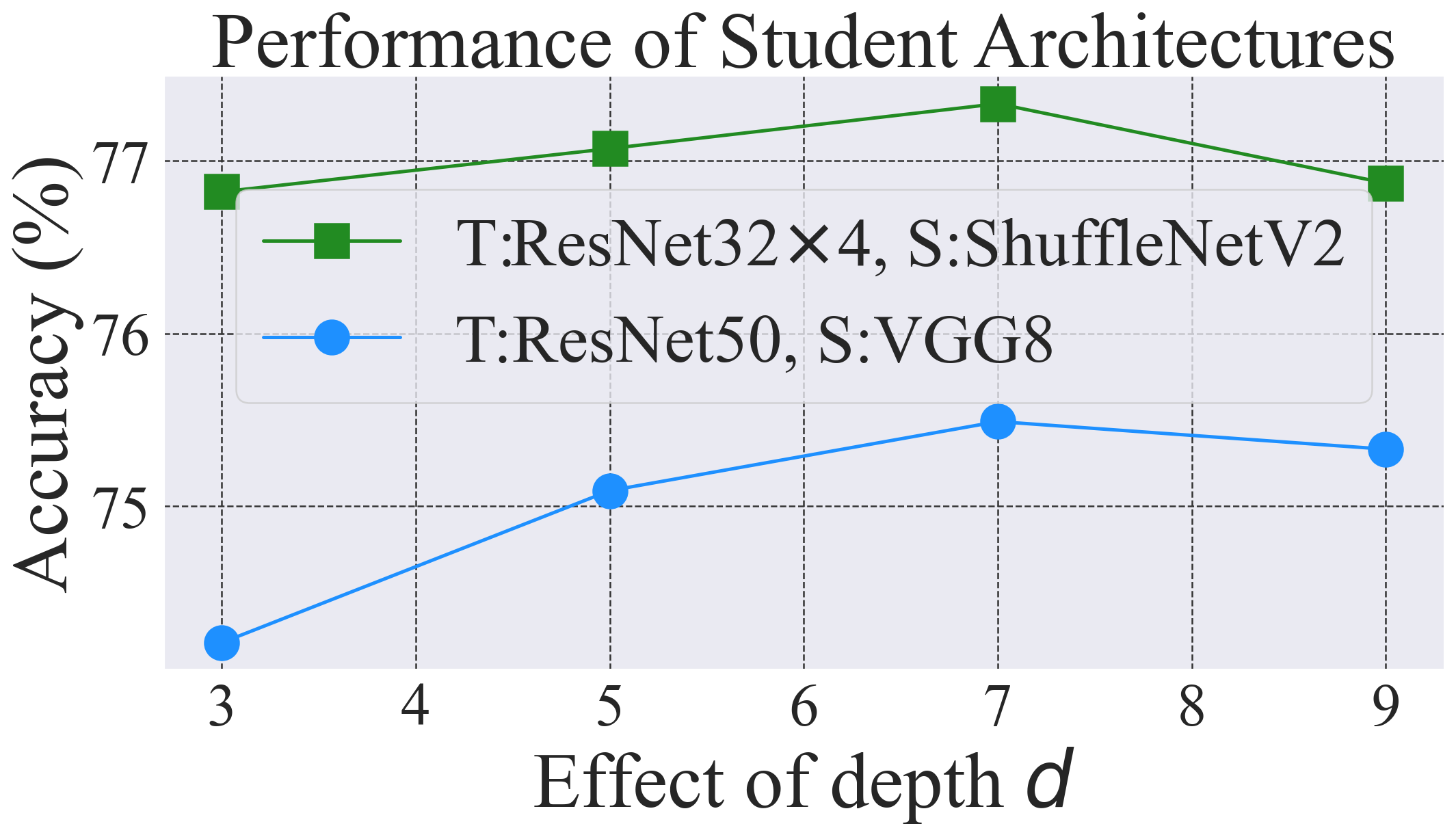}
\vspace{-18pt}
        \caption{ Heterogeneous architectures}
    \end{subfigure}
    \vspace{-19pt}
\caption{Impact of the depth  $d$ on the performance of the student on CIFAR-100.}
\label{depth}
\end{figure}

\textbf{ Impact of the recursion depth $d$.}
We studied the impact of the recursion depth $d$ on student performance. When the depth $d$ is small, the limited number of category pairs hinders the student from capturing sufficient inter-class distinctions. Conversely, a large depth $d$ may introduce significant noise, interfere with the learning process, and lead to information redundancy due to the large distances between categories.~Thus, excessive recursion depth increases optimization complexity without improving performance.~Our findings indicate that a moderate recursion depth, particularly the depth $d = 7$ in Fig.~\ref{depth}, achieves the optimal performance across homogeneous and heterogeneous architectures.

\textbf{ Impact of the ADW.}
We conducted an ablation study to evaluate the impact of the component $\Omega_{ADW}$ on student performance.~As shown in Table \ref{Impact of the ADW}, using $\mathcal{L}$ and $\mathcal{L}_{LLKI}$, the performance is quite favorable compared to training the student from scratch.~After incorporating  $\Omega_{ADW}$, the performance improves further.~The results show that the component $\Omega_{ADW}$, which adjusts category pair weights, can enhance discriminability and effectively improve the  performance of the student.
\begin{table}[!ht]
\centering
\renewcommand{\arraystretch}{1.0}
\resizebox{\linewidth}{!}{
\begin{tabular}{ccc|cccccc}
\tophline
\multicolumn{3}{c|}{$\mathcal{L}_{Local}$} &VGG13& VGG13 & ResNet50  & ResNet32$\times$4 \\
\cmidrule(lr){1-3} 
\multirow{1}*{$\mathcal{L}$} & \multirow{1}*{$\Omega_{ADW}$} & \multirow{1}*{$\mathcal{L}_{LLKI}$}  & VGG8& MobileNetV2 & MobileNetV2 & ShffleNetV2 \\
\midhline
- & - & - & 70.50 & 64.60 & 64.60 & 71.82 \\
\checkmark & & \checkmark & 74.25 (\textcolor{deepgreen}{+3.75}) & 68.95 (\textcolor{deepgreen}{+4.35}) & 69.23 (\textcolor{deepgreen}{+4.63})  & 76.92 (\textcolor{deepgreen}{+5.10}) \\
\checkmark & \checkmark & \checkmark & \textbf{74.46} (\textcolor{deepgreen}{+3.96}) & \textbf{69.27} (\textcolor{deepgreen}{+4.67}) & \textbf{69.52} (\textcolor{deepgreen}{+4.92}) & \textbf{77.07} (\textcolor{deepgreen}{+5.25}) \\
\bottomline
\end{tabular}
    }
\caption{\footnotesize{Impact of ADW on student’s performance on CIFAR100.}}
\label{Impact of the ADW}
\vspace{-13pt}
\end{table}
\subsection{Generalization Performance Evaluation}
\vspace{-3pt}
To validate the generalization ability of our method, we extended the task to person Re-ID.~As presented in Table~\ref{reid}, the results indicate that our method exhibits superior performance compared to other distillation methods, such as AT and DKD, confirming its effectiveness.
\vspace{-10pt}
\begin{table}[!h]
\centering
\resizebox{\linewidth}{!}{
\begin{tabular}{l|c|c|c|c}
\tophline
Methods & Rank@1 & Rank@5 & Rank@10 & mAP \\
\hline ResNet50(teacher) & 88.03 & 95.01 & 96.67 & 71.77 \\
ResNet18(student) & 85.04 & 94.06 & 96.23 & 65.30 \\
\hline
AT~\cite{komodakis2017paying} & 86.71 & 94.88 & 96.89 & 68.42 \\
KD~\cite{hinton2015distilling} & 88.21 & 94.88 & 96.80 & 71.79 \\
DKD~\cite{zhao2022decoupled} & 87.50 & 94.74 & 96.71 & 71.83 \\
\hline LDRLD & $\textbf{88.31}$ & $\textbf{95.05}$ & $\textbf{97.03}$ & $\textbf{72.13}$ \\
 $\Delta$ & ${\color{blue}+3.27}$ & ${\color{blue}+0.99}$ & ${\color{blue}+0.80}$ & ${\color{blue}+6.83}$ \\
\bottomline
\end{tabular}
}
\vspace{-0.2cm}
\caption{Comparison with different methods on Market-1501.}
\label{reid}
\end{table}

\subsection{Visualizations}
\vspace{-0pt}
We evaluate our experimental results using Grad-CAM~\cite{selvaraju2017grad} and Correlation Matrices for visualization and the latter is provided in the \textit{Supplementary Material}.

\begin{figure}[!h]
\centering
\begin{minipage}{0.14\linewidth}
     \vspace{3pt}  
    \begin{overpic}[width=\textwidth]{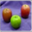}
        \put(8.5,120){\color{black} \footnotesize Orignal} 
    \end{overpic}
     \centerline{\includegraphics[width=\textwidth]{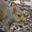}}
     
     \centerline{\includegraphics[width=\textwidth]{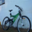}}
     \centerline{\includegraphics[width=\textwidth]{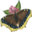}}
     
     \centerline{\footnotesize  (a) Input}
 \end{minipage}
    \hspace{0.1cm} 
         \begin{minipage}{0.0000000001\textwidth}
            \begin{tikzpicture}
                \draw[dashed] (0.1,0) -- (0.1,5.1);
            \end{tikzpicture}
            \end{minipage}
    \hspace{0.1cm} 
 \begin{minipage}{0.14\linewidth}
     \vspace{3pt}  
     \begin{overpic}[width=\textwidth]{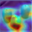}
        \put(-8,120){\color{black} \footnotesize ResNet32×4}
    \end{overpic}
     \centerline{\includegraphics[width=\textwidth]{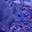}}
     \centerline{\includegraphics[width=\textwidth]{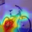}}
     \centerline{\includegraphics[width=\textwidth]{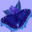}}
     \centerline{\footnotesize  (b) KD}
 \end{minipage}
 \hspace{0.1cm} 
    \begin{minipage}{0.14\linewidth}
     \vspace{3pt}
     \begin{overpic}[width=\textwidth]{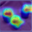}
        \put(-10,120){\color{black} \footnotesize {/ResNet8×4}}
    \end{overpic}
     \centerline{\includegraphics[width=\textwidth]{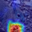}}
     
     \centerline{\includegraphics[width=\textwidth]{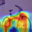}}
     \centerline{\includegraphics[width=\textwidth]{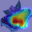}}
     
     \centerline{\footnotesize  (c) LDRLD}
 \end{minipage}
        \hspace{0.1cm} 
         \begin{minipage}{0.0000000001\textwidth}
            \begin{tikzpicture}
                \draw[dashed] (0.1,0) -- (0.1,5.1);
            \end{tikzpicture}
            \end{minipage}
        \hspace{0.1cm} 
    \begin{minipage}{0.14\linewidth}
     \vspace{3pt}
     \begin{overpic}[width=\textwidth]{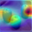}
        \put(-8,120){\color{black} \footnotesize ResNet32×4}
    \end{overpic}
     \centerline{\includegraphics[width=\textwidth]{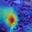}}
     
  \centerline{\includegraphics[width=\textwidth]{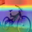}}
     \centerline{\includegraphics[width=\textwidth]{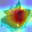}}
     
     \centerline{\footnotesize (d) KD}
 \end{minipage}
 \hspace{0.1cm} 
    \begin{minipage}{0.14\linewidth}
     \vspace{3pt}
     \begin{overpic}[width=\textwidth]{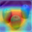}
        \put(-10,120){\color{black} \footnotesize {/ShuffleV2}}
    \end{overpic}
     
     \centerline{\includegraphics[width=\textwidth]{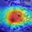}}
     
     \centerline{\includegraphics[width=\textwidth]{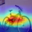}}
     
\centerline{\includegraphics[width=\textwidth]{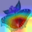}}
     
     \centerline{\footnotesize (e) LDRLD}
 \end{minipage}
	\caption{Feature map visualization of the student's penultimate layers on CIFAR-100 dataset using vanilla KD and LDRLD.}
	\label{cam}
\end{figure}
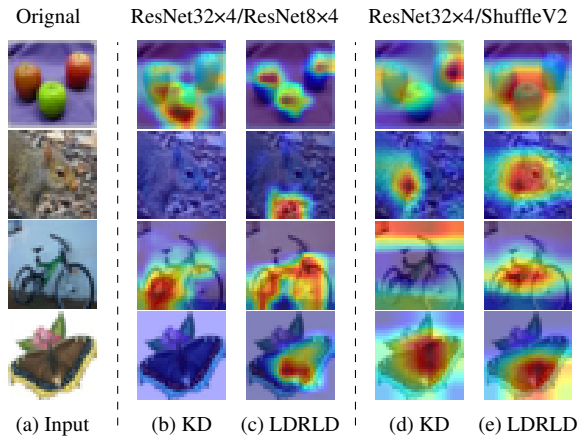

\textbf{Visualization of Grad-CAM.} To demonstrate the effectiveness of our proposed method, we use Grad-CAM to visualize the feature map. As shown in Figs.~\ref{cam} (c) and (e), our method accurately identifies target objects by focusing on the detailed features of primary categories. In contrast, Figs.~\ref{cam} (b) and (d) reveal that vanilla KD tends to focus on non-critical regions, causing it to overlook crucial information,  and leading to performance degradation.~Instead, Our method’s targeted attention can retain important details, thereby enhancing overall
performance.

\section{Conclusion}
\vspace{-3pt}
We propose the Local Dense Relational Logit Distillation method and further incorporate the Adaptive Decay Weight strategy, including Inverse Rank Weighting and Exponential Rank Decay, to effectively capture fine-grained features within logit knowledge.~Experimental results demonstrate that our method outperforms state-of-the-art logit-based KD methods on the CIFAR-100, Tiny-ImageNet, MSCOCO-2017, and ImageNet-1K datasets, validating its effectiveness.\\
\textbf{Limitations and future work.}~Despite these achievements, our method still requires manual selection of the recursion depth $d$. Since the optimal choice of 
$d$ depends on the number of classes and the complexity of the task, it is necessary to adjust 
$d$ accordingly. Therefore, future work will explore adaptive approaches to address this limitation.
\section*{Acknowledgement}
The work was supported in part by the National Natural Science Foundation of China (U21A20487, 62273082), in part by the Guangdong Technology Project (2023TX07Z126, 2022B1515120067), in part by the Shenzhen Technology Project (JCYJ20220818101206014, JCYJ20220818101211025, GJHZ20240218112504008), in part by Yunnan Science \& Technology Project (202305AF150152, 202302AD080008).
{
    \small
    \bibliographystyle{ieeenat_fullname}
    \bibliography{main}
}

\clearpage
\setcounter{page}{1}
\maketitlesupplementary
\setcounter{section}{0}
\section{Optimization Objective}
The algorithm for implementing the LDRLD is provided in Algorithm~\ref{alg}, detailing the essential steps.
\begin{algorithm}[!ht]
\caption{Pseudo code for LDRLD}
\begin{algorithmic}[1]
\Require 
    $\mathcal{D}$: training dataset; 
    $T_{\text{net}}$: pre-trained teacher network; 
    $S_{\text{net}}$: student network with parameters $\theta$; 
    $\eta$: learning rate
\Ensure 
Trained parameters $\theta$ of the student network $S_{\text{net}}$
\State Load pre-trained teacher network $T_{\text{net}}$
\Repeat
    \State Randomly select a mini-batch $\mathcal{B}$ from $\mathcal{D}$
    \For{each $(\mathbf{x}_i, \mathbf{y}_i) \in \mathcal{B}$}
        \State $\mathbf{Z}^{t} \leftarrow T_{\text{net}}(\mathbf{x}_i)$ \Comment{Compute teacher's output for sample $i$ and teacher's parameters are fixed}.
        \State $\mathbf{Z}^{s} \leftarrow S_{\text{net}}(\mathbf{x}_i)$ \Comment{Compute student's output for sample $i$}.
        \State Compute the task loss $\mathcal{L}_{\text{Task}}$ (i.e., cross entropy). 
        \State Compute the local logit relational knowledge $\mathcal{L}^w$ using Equation~(\ref{local}).
        \State Compute the remaining non-target knowledge $\mathcal{L}_{\text{RNTK}}$ using Equation~(\ref{RNTK}).
        \State Compute the total loss $\mathcal{L}_{\text{LDRLD}}$ by combining the task loss $\mathcal{L}_{\text{Task}}$, $\mathcal{L}^w$, and $\mathcal{L}_{\text{RNTK}}$, as described in Equation~(\ref{total_loss}).
    \EndFor
    \State $\mathcal{L} \leftarrow \dfrac{1}{|\mathcal{B}|} \sum_{i} \mathcal{L}_{\text{LDRLD}}$
    \State Update $\theta \leftarrow \theta - \eta \left( \nabla_{\theta} \mathcal{L} \right)$
\Until{Convergence criterion is maximum iterations reached}
\State \Return $\theta$
\end{algorithmic}
\label{alg}
\end{algorithm}
\vspace{-12pt}
\section{Experiments}
\subsection{Experimental Setups}
In our experiments, we evaluate our method using the following five classical datasets.

\textbf{CIFAR-100}~\cite{krizhevsky2009learning} comprises 100 classes, each image with a resolution of 32$\times$32 pixels.~The dataset contains 50,000 training images and 10,000 validation images.

\textbf{ImageNet-1K (ILSVRC2012)}~\cite{deng2009imagenet} is a comprehensive dataset comprising 1,000 classes. The dataset contains 1.2 million training images and 50,000 validation images. 

\textbf{Tiny-ImageNet}~\cite{le2015tiny} is a streamlined version of the ImageNet-1K dataset. The dataset includes 200 classes, with images of 64$\times$64 pixels resolution.~It comprises 100,000 training images and 10,000 validation images.

\textbf{Market-1501}~\cite{zheng2015scalable} is a benchmark dataset for person re-identification (Re-ID), containing 1,501 unique identities and 32,668 images. It includes a query set of 3,368 images covering 751 identities (4–6 images each) and a gallery set of 30,368 images from 12 camera viewpoints.

\textbf{MS-COCO2017}~\cite{lin2014microsoft} is a widely used large-scale dataset for object detection, consisting of 80 categories, with 118,000 images in the train2017 split and 5,000 images in the val2017 split.


\subsection{Implementations: Comparison with state-of-the-art KD Methods.}

In this work, we compare the proposed method with classical benchmarks across several datasets, including CIFAR-100, ImageNet-1K, Market-1501, Tiny-ImageNet, and COCO2017. The comparison focuses on two main categories of KD methods. We evaluate the following feature-based KD methods:
FitNets~\cite{romero2014fitnets}, 
SP~\cite{tung2019similarity}, CC~\cite{peng2019correlation}, VID~\cite{ahn2019variational}, 
AT~\cite{komodakis2017paying}, 
OFD~\cite{heo2019comprehensive},
PKT~\cite{passalis2018learning}, 
NST~\cite{huang2017like},
RKD~\cite{park2019relational}, 
FT~\cite{kim2018paraphrasing},CRD~\cite{tian2019contrastive}, 
SAKD~\cite{song2022spot},
ReviewKD~\cite{chen2021distilling}, NORM~\cite{liu2023norm},FCFD~\cite{liu2023functionconsistent}, DiffKD~\cite{huang2024knowledge},
and CAT-KD~\cite{guo2023class}. We also evaluate the following logit-based KD methods such as:  KD~\cite{hinton2015distilling}, DKD~\cite{zhao2022decoupled}, NKD~\cite{yang2023knowledge}, CTKD~\cite{li2023curriculum}, LCKA~\cite{Zhou2024RethinkingCK}, WTTM~\cite{zheng2024knowledge}, 
IPWD~\cite{niu2022respecting},
SDD~\cite{luo2024scale}, and LSKD~\cite{Sun2024Logit}.

\begin{table*}[!h]
\centering
\setlength\tabcolsep{8.5pt}
\scalebox{0.75}{
\begin{tabular}{ccccccc}
\tophline
\begin{tabular}{c} 
Teacher \\ Student
\end{tabular} & 
\begin{tabular}{c} 
VGG13 \\ MobileNetV2
\end{tabular} & 
\begin{tabular}{c} 
ResNet50 \\ MobileNetV2
\end{tabular} & 
\begin{tabular}{c} 
ResNet50 \\ VGG8
\end{tabular} & 
\begin{tabular}{c} 
ResNet32$\times$4 \\ ShuffleNetV1
\end{tabular} & 
\begin{tabular}{c} 
ResNet32$\times$4 \\ ShuffleNetV2
\end{tabular} & 
\begin{tabular}{c} 
WRN-40-2 \\ ShuffleNetV1
\end{tabular}\\
\hline
LDRLD & $\alpha=7.0, \beta=4.0$ & $\alpha=11.0, \beta=7.0$ & $\alpha=9.5, \beta=3.5$ & $\alpha=8.0, \beta=8.0$ & $\alpha=9.5, \beta=7.0$ & $\alpha=11.0, \beta=7.0$ \\
\tophline
\end{tabular}
}
\vspace{-6pt}
\caption{Hyperparameters for heterogeneous architecture distillation on CIFAR-100 dataset.}
\label{table:teacher_student_diff}
\end{table*}

\begin{table*}[!h]
\centering
\setlength\tabcolsep{7.0pt}
\scalebox{0.7}{
\begin{tabular}{cccccccc}
\tophline
\begin{tabular}{c} 
Teacher \\ Student
\end{tabular} & 
\begin{tabular}{c} 
WRN-40-2 \\ WRN-16-2
\end{tabular} & 
\begin{tabular}{c} 
WRN-40-2 \\ WRN-40-1
\end{tabular} & 
\begin{tabular}{c} 
ResNet56 \\ ResNet20
\end{tabular} & 
\begin{tabular}{c} 
ResNet110 \\ ResNet20
\end{tabular} & 
\begin{tabular}{c} 
ResNet110 \\ ResNet32
\end{tabular} & 
\begin{tabular}{c} 
ResNet32$\times$4 \\ ResNet8$\times$4
\end{tabular} & 
\begin{tabular}{c} 
VGG13 \\ VGG8
\end{tabular}\\
\hline
LDRLD & $\alpha=11.5, \beta=7.0$ & $\alpha=9.0, \beta=4.0$ & $\alpha=9.5, \beta=1.0$ & $\alpha=6.5, \beta=1.0$ & $\alpha=8.0, \beta=1.0$ & $\alpha=10.5, \beta=7.0$ & $\alpha=11.0, \beta=8.5$ \\
\tophline
\end{tabular}
}
\vspace{-6pt}
\caption{Hyperparameters for homogeneous architecture distillation on CIFAR-100 dataset.}
\label{table:teacher_student_same}
\end{table*}

\begin{table*}[!h]
\centering
\setlength\tabcolsep{9.0pt}
\scalebox{0.8}{
\begin{tabular}{cccccc}
\tophline
\begin{tabular}{c} 
Teacher \\ Student
\end{tabular}  & 
\begin{tabular}{c} 
WRN-40-2 \\ WRN-16-2
\end{tabular} & 
\begin{tabular}{c} 
ResNet56 \\ ResNet20
\end{tabular} & 
\begin{tabular}{c} 
ResNet110 \\ ResNet20
\end{tabular} & 
\begin{tabular}{c} 
VGG13 \\ MobileNetV2
\end{tabular} & 
\begin{tabular}{c} 
VGG13 \\ VGG8
\end{tabular}\\
\hline
LDRLD & $\alpha=8.5, \beta=5.0$ & $\alpha=8.0, \beta=4.0$ & $\alpha=8.5, \beta=5.0$ & $\alpha=6.0, \beta=4.0$ & $\alpha=8.5, \beta=4.5$ \\
\tophline
\end{tabular}
}
\vspace{-6pt}
\caption{Hyperparameters for homogeneous and heterogeneous architectures distillation on Tiny-ImageNet dataset.}
\label{table:tiny_teacher_student_same}
\end{table*}

\begin{table}[!ht]
\vspace{-6pt}
\centering 
\setlength\tabcolsep{10.0pt}
\scalebox{0.9}{
\begin{tabular}{c|c|cc}
\tophline
\multirow{2}*{Distillation}&Teacher & ResNet34& ResNet50 \\ 
& Student & ResNet18 & MobileNetV1 \\
\hline
\multirow{2}*{LDRLD}&$\alpha$  & 7.0 & 5.0 \\
&$\beta$  & 0.025 & 2.0 \\
\tophline
\end{tabular}
}
\vspace{-6pt}
\caption{Hyperparameters for homogeneous and heterogeneous architectures distillation on the ImageNet-1K dataset.}
\label{Hyper_imagenet}
\vspace{-12pt}
\end{table}

\subsection{Implementations: Selection of Model Architectures.}
To thoroughly evaluate the effectiveness of our proposed method, we employed several classical network architectures for image classification, including ResNet~\cite{he2016deep} (with its variants), VGG~\cite{simonyan2014very}, MobileNetV1~\cite{howard2017mobilenets}, WideResNet~\cite{2016wide} (WRN), ShuffleNetV1~\cite{zhang2018shufflenet}, ShuffleNetV2~\cite{ma2018shufflenet}, and MobileNetV2~\cite{sandler2018mobilenetv2}. In our experimental framework, we used a variety of comparative approaches to assess the performance of different teacher-student pairs. We conducted experiments on the CIFAR-100, Tiny-ImageNet, and ImageNet-1K datasets using student networks with identical and different architectures.~Additionally, we performed experiments with teacher-student pairs using identical architectures on the Market-1501 dataset. These experiments validated the effectiveness and consistency of our method across various datasets and network architectures.
\subsection{Implementations: Training Details.}\textbf{For both the Tiny-ImageNet and CIFAR-100 datasets:} We train the models for 240 epochs using the Stochastic Gradient Descent optimizer with a momentum of 0.9 and a weight decay of $5 \times 10^{-4}$. The learning rate is reduced by a factor of 10 at the 150th, 180th, and 210th epochs. Data augmentation is performed using random horizontal flipping. We set the recursion depth to the default value of $d=7$. The temperature coefficient $\tau$ is set to 4.0 across all experiments. The specific training details are as follows:
\begin{itemize}
\item (1) For the CIFAR-100 dataset, we use a batch size of 64. The learning rate is set to 0.05 for all architectures except MobileNetV2~\cite{sandler2018mobilenetv2} and ShuffleNet~\cite{ma2018shufflenet,zhang2018shufflenet}, for which it is set to 0.01. The hyperparameters are provided in Table~\ref{table:teacher_student_diff} and Table~\ref{table:teacher_student_same}, where the recursion depth is set to $d=9$ for ResNet32$\times$4 vs. ShuffleNetV1 and to $d=7$ for all other models.~All models are trained with a linear warm-up for 20 epochs. See CRD~\cite{tian2019contrastive} for detailed training details.

\item (2) For the Tiny-ImageNet dataset, we use a batch size of 128. The learning rate is set to 0.1 for all models except MobileNetV2~\cite{sandler2018mobilenetv2}, for which it is set to 0.02. The weighting coefficient for the cross-entropy is set to 1.0. All models are trained with a linear warm-up for 20 epochs. For more details on the hyperparameters, refer to Table~\ref{table:tiny_teacher_student_same}.
\end{itemize}
 \textbf{For training on the ImageNet-1K dataset:} We train the models for 100 epochs using the Stochastic Gradient Descent optimizer with a weight decay of $1 \times 10^{-4}$. The batch size of 256 is used for all models, and the initial learning rate is set to 0.1. A linear warm-up is applied during the first 10 epochs. The learning rate is then reduced by a factor of 10 at the 30th, 60th, and 90th epochs. The loss of cross-entropy is weighted by a coefficient of 1.0, and the temperature coefficient $\tau$ is set to 2.0. For model pairing, we use ResNet34 as the teacher and ResNet18 as the student when using the same architecture. For different architectures, ResNet50 serves as the teacher and MobileNetV1 as the student. For further details on the hyperparameters, please refer to the Table~\ref{Hyper_imagenet}.

 \noindent\textbf{For training on the Matket-1501 dataset:} We employ ResNet50 as the backbone to extract features for the teacher network and ResNet18 for the student network. To evaluate the performance of person Re-identification (Re-ID), we use two commonly applied metrics: Cumulative Matching Characteristics (CMC) and mean Average Precision (mAP). The CMC-$k$ metric (i.e., Rank-$k$ matching accuracy) measures the probability that a correct match appears in the top-$k$ ranked retrieval results. CMC performs well when there is only one ground truth per query because it focuses solely on the first correct match among the top-$k$ results during the evaluation. For more detailed explanations of these metrics, please refer to~\cite{Post,Re-Identification}.
\vspace{-6pt}

\section{Ablation Study}
\subsection{Hyperparameter Sensitivity}
\noindent\textbf{ Impact of $\alpha$ and $\beta$ on CIFAR-100.} The results presented in Tables \ref{cifar100-ha} and \ref{cifar100-hb} show the accuracy of student (\%) with different values of $\alpha$ and $\beta$ across various architectures. We evaluated the impact of specific parameter values on student performance while keeping other parameters fixed. As shown in Table~\ref{cifar100-ha}, for homogeneous pairs of teacher-student, the experimental results show optimal student performance with $\alpha$ = 9.5 and $\beta$ = 1.0. Similarly, as shown in Table~\ref{cifar100-hb}, for heterogeneous pairs of teacher-student, the experimental results demonstrate optimal student performance with $\alpha$ = 9.5 and $\beta$ = 7.0.~These results indicate that the optimal parameter settings differ across architectures.
\begin{table}[!ht]
\centering
\setlength\tabcolsep{4.9pt}
\renewcommand{\arraystretch}{1.2} 
\scalebox{0.72}{
\begin{tabular}{lccccccccc}
\tophline
$\alpha$ & 6.0 & 6.5 & 7.0 & 7.5 & 8.0&8.5 &9.0&\textbf{9.5}&10.0\\
Acc & 71.87&72.03& 71.95& 71.94& 71.96&71.94&71.90&\textbf{72.20}&71.85  \\
\hline
$\beta$ & 0.75 & \textbf{1.0} & 1.25 & 1.5 & 2.0&2.25 &2.5&3.0&4.0\\
Acc & 72.01&\textbf{72.20}& 71.62& 72.04&71.81& 71.92&72.05&71.86 &71.78 \\
\tophline
\end{tabular}
}
\caption{Impact of $\alpha$ and $\beta$ on student's performance(\%) on CIFAR-100 using ResNet56 as teacher and ResNet20 as student.}
\label{cifar100-ha}
\end{table}
\begin{table}[!ht]
\centering
\setlength\tabcolsep{4.9pt}
\renewcommand{\arraystretch}{1.2} 
\scalebox{0.72}{
\begin{tabular}{lccccccccc}
\tophline
$\alpha$ & 6.5 & 7.0 & 7.5 & 8.0&8.5 &9.0&\textbf{9.5}&10.0& 10.5 \\
Acc & 77.15&77.10& 76.99& 76.98& 77.14&77.30&\textbf{77.33}&76.93&76.87  \\
\hline
$\beta$ & 5.0 & 5.5 & 6.0 & 6.5 & \textbf{7.0}&7.5 &8.0&8.5&9.0\\
Acc & 77.01&77.03& 76.65& 76.92&\textbf{77.33}& 77.04&76.83&77.02 &76.75 \\
\tophline
\end{tabular}
}
\caption{Impact of $\alpha$ and $\beta$ on student's performance(\%) on CIFAR-100 using ResNet32$\times$4 as teacher and ShuffleNetV2 as student.}
\label{cifar100-hb}
\end{table}

\noindent\textbf{Impact of $\alpha$ and $\beta$ on Tiny-ImageNet.} The results presented in Tables~\ref{tinyimagenet-ha} and \ref{tinyimagenet-hb} show the student accuracies (\%) across various architectures for different values of $\alpha$ and $\beta$, while holding other parameters fixed. Specifically, for the teacher-student pair WRN-40-2 and WRN-16-2 (Table~\ref{tinyimagenet-ha}), the best performance is achieved with $\alpha = 8.5$ and $\beta = 5.0$. Similarly, for the teacher-student pair VGG13 and MobileNetV2 (Table~\ref{tinyimagenet-hb}), the optimal performance occurs at $\alpha = 6.0$ and $\beta = 4.0$. These results demonstrate that the optimal parameter settings vary across different architectures. For more detailed parameter settings, see Table~\ref{table:tiny_teacher_student_same}.
\begin{table}[!ht]
\centering
\setlength\tabcolsep{7.0pt}
\renewcommand{\arraystretch}{1.2} 
\scalebox{0.72}{
\begin{tabular}{lcccccccc}
\tophline
$\alpha$  & 5.0 & 5.5 & 6.0&6.5 &7.0&7.5&8.0&\textbf{8.5}\\
Acc & 60.65&60.36& 60.58& 60.26& 60.62&60.63&60.49&\textbf{60.67}\\
\hline
$\beta$ &  4.5& \textbf{5.0}& 5.5 & 6.0&6.5 &7.0&7.5&8.0\\
Acc & 60.58& \textbf{60.67}&60.65& 60.62& 60.29&60.40& 60.48&60.38 \\
\tophline
\end{tabular}
}
\caption{Effect of $\alpha$ and $\beta$ on student's performance on Tiny-Imagenet using WRN-40-2 as teacher and WRN-16-2 as student.}
\label{tinyimagenet-ha}
\end{table}
\begin{table}[!ht]
\centering
\setlength\tabcolsep{7.0pt}
\renewcommand{\arraystretch}{1.2} 
\scalebox{0.72}{
\begin{tabular}{lcccccccc}
\tophline
$\alpha$  & 5.0 & 5.5 & \textbf{6.0}&6.5 &7.0&7.5&8.0&8.5\\
Acc & 60.24&60.00& \textbf{60.31}& 60.29& 60.27&60.19&60.02&59.98\\
\hline
$\beta$ &  1.0& 2.0& 3.0 & 3.5&\textbf{4.0} &4.5&5.0&5.5\\
Acc & 59.73& 60.05&60.06& 60.18& \textbf{60.31}&60.05& 59.77&60.08 \\
\tophline
\end{tabular}
}
\caption{Effect of $\alpha$ and $\beta$ on student's performance on Tiny-Imagenet using VGG13 as teacher and MobileNetV2 as student.}
\label{tinyimagenet-hb}
\vspace{-10pt}
\end{table}
\subsection{Ablation and explanation of ADW}~(1) Results shown in Table~\ref{Ablation of two losses} demonstrate that each loss improves performance while combining both losses achieves the best outcome.~(2) $\mathcal{L}_{Local}$ still outperforms the baseline without ADW in Table~\ref{Impact of the ADW}.~However, without ADW, using uniform weights for logit pairs fails to mitigate class confusion caused by semantic gaps. In contrast, ADW can enhance fine-grained logit discrimination and improve performance in Table~\ref{Impact of the ADW}. 
\begin{table}[!ht]
\vspace{-7.8pt}
\centering
\renewcommand{\arraystretch}{1.4}
\resizebox{\linewidth}{!}{
\begin{tabular}{cc|cccccc}
\tophline
\multirow{2}*{$\mathcal{L}_{Local}$}&\multirow{2}*{$\mathcal{L}_{RNTK}$}  & ResNet32$\times$4 & WRN-40-2 &ResNet32$\times$4 & WRN-40-2 \\
&  &ResNet8$\times$4& WRN-40-1 & ShuffleNetV2  & ShuffleNetV1 \\
\midhline
 -& - & 72.50&  71.98 & 71.82  & 70.50 \\
 \checkmark  && 76.55 (\textcolor{deepgreen}{+4.05})& 74.45 (\textcolor{deepgreen}{+2.47}) & 76.23 (\textcolor{deepgreen}{+4.41})  & 76.11 (\textcolor{deepgreen}{+5.61}) \\
  &\checkmark& 75.78 (\textcolor{deepgreen}{+3.28})& 74.19 (\textcolor{deepgreen}{+2.21}) & 76.30 (\textcolor{deepgreen}{+4.48})  & 76.51 (\textcolor{deepgreen}{+6.01}) \\
 \checkmark &\checkmark& \textbf{77.20} (\textcolor{deepgreen}{+4.70})& \textbf{74.98} (\textcolor{deepgreen}{+3.00}) & \textbf{77.33} (\textcolor{deepgreen}{+5.51})  & \textbf{77.09} (\textcolor{deepgreen}{+6.59}) \\ 
\bottomline

\end{tabular}
}
\vspace{-0.1cm}
\caption{\footnotesize{Ablation of loss on student’s performance on CIFAR100.}}
\label{Ablation of two losses}
\end{table}

\subsection{Training Efficiency}
To evaluate the training cost of our method, we compare it with classical KD techniques by reporting the average runtime per epoch, as shown in Table \ref{table:training efficiency}. By incorporating local logit decoupling and combination mechanisms, our method maintains high training efficiency without additional storage and with only a slight increase in runtime.~Although LDRLD requires additional training time, it requires GPU memory very similar to logit-based KD methods such as KD and DKD without increasing memory consumption. Furthermore, compared to feature-based KD methods such as CRD and ReviewKD, our approach achieves more efficient training.
\begin{table}[!ht]
\centering
\resizebox{0.99\linewidth}{!}{
\begin{tabular}{ccccccc}
\tophline
Distillation Manner& \multicolumn{2}{c}{Feature-based} & \multicolumn{3}{c}{Logit-based} \\ \cmidrule(lr){2-3} \cmidrule(lr){4-6}
  Method&  CRD & ReviewKD  KD &  DKD &  LDRLD \\
 \midhline
\multirow{1}{*}{Time(s)} & 17.86 & 23.44 & 11.89 & 12.04 & 13.65 \\ 
\multirow{1}{*}{GPU memory (M)} & 3064 & 3148 & 2052 & 2052 &  2052 \\ 
\multirow{1}{*}{Accuracy(\%)} & 75.51 & 75.63 & 73.33 & 76.32 & 77.20 \\ 
\bottomline
\end{tabular}
}
\caption{We assess the average training time (per epoch) and test accuracy (\%) using a GeForce RTX 3090 on the CIFAR-100 dataset for ResNet32$\times$4 (teacher) and ResNet8$\times$4 (student).}
\label{table:training efficiency}
\vspace{-0.5cm}
\end{table}
\section{Exploratory  Experiments}
\subsection{Differences between NCKD and RNTK}
DKD decoupling results in NCKD~\cite{zhao2022decoupled}, which excludes only the target class, whereas RNTK excludes the top-1 to top-$d$ most confident classes.~The latter strategy avoids suppressing high-confidence categories, thereby reducing dependence on common categories, improving the ability to recognize unseen data, and promoting the activation of low-confidence classes.~The experimental results demonstrate that the student with RNTK achieves the best performance when $d$=7, as shown in Table \ref{tiny-ImageNet-CUB1}, outperforming NCKD and supporting this argument.
\vspace{-0.5mm}
\begin{table}[!ht]
\centering
\setlength\tabcolsep{6.5pt}
\scalebox{0.9}{
\begin{tabular}{c|c|c|c|c|c}
\tophline
Model & NCKD & \multicolumn{4}{c}{RNTK} \\
\midhline
T:VGG13 & - & d = 3 & d = 5 & d = 7 & d = 9 \\
\cline{2-6}
S:VGG8 & 74.15 & 73.76 & 74.19 & \textbf{74.28} & 74.26 \\
\midhline
T:ResNet32$\times$4 & - &d = 3 & d = 5 & d = 7 & d = 9 \\
\cline{2-6}
S:ResNet8$\times$4 & 75.42 & 75.34 & 75.73 & \textbf{75.78} & 75.61 \\
\bottomline
\end{tabular}}
\vspace{-3pt}
\caption{A comparison of the top-1 performance of students trained with NCKD and RNTK on the CIFAR-100 dataset.}
\label{tiny-ImageNet-CUB1}
\end{table}

\subsection{Influence of incorrect teacher predictions}
 We consider the case that the teacher model is poorly calibrated, biased, or makes incorrect predictions, and these issues could propagate to the student model during distillation. In order to deal with these problems, ~we introduce noisy labels with a noise rate of 0.1 \cite{han2018co}, and the experimental results in Table \ref{table:noisy-left} show that even if the teacher is biased or inaccurate, the LDRLD can still effectively transfer knowledge and perform robust. This shows, indicating that the recursive combination strategy may correct the influence of incorrect logits,.

\begin{table}[!ht]
\centering
\renewcommand{\arraystretch}{1.0}  
\resizebox{\linewidth}{!}{
\begin{tabular}{c|cccc}
\tophline
Teacher&ResNet110&ResNet110&ResNet32$\times$4&VGG13\\
Student&ResNet20&ResNet32&ResNet8$\times$4&VGG8\\
\midhline
T:Accuracy & 74.31 & 74.31 & 79.42 & 74.64 \\
S:Accuracy & 69.06 & 71.14 & 72.50 & 70.36  \\
\midhline
DKD & 70.91 & 74.11 & 76.32 & 74.64 \\
DKD(0.1) & 71.32 & 73.81 & 76.12 & 74.32  \\
$\Delta$&{\color{blue}+0.41}&{\color{red}-0.30}&{\color{red}-0.20}&{\color{red}-0.32}\\
\midhline
LDRLD & 71.98 & 74.16 & 77.20 & 75.06 \\
LDRLD(0.1) & 72.08 & 74.13 & 77.32 & 74.91  \\
$\Delta$&{\color{blue}+0.10}&{\color{red}-0.03}&{\color{blue}+0.12}&{\color{red}-0.15}\\
\bottomline
\end{tabular}
}
\caption{The top-1 performance of students with noisy labels on the CIFAR-100 datasets.}
\label{table:noisy-left}
\vspace{-18pt}
\end{table}

\subsection{Predictions of different teachers}
To address knowledge capacity gap, we designed different teachers for the same student, as shown in
Table \ref{table:noisy-right}. As the accuracy of the teacher increases, LDRLD generally delivers logit knowledge more accurately than DKD, thereby enabling the student's performance to more closely approximate that of the teacher, thanks to the accurate leverage of inter-class information by logit combinations.
\begin{table}[!ht]
\centering
\renewcommand{\arraystretch}{1.2}  
\resizebox{\linewidth}{!}{
\begin{tabular}{c|cccc}
\tophline
Teacher&ResNet32&ResNet44&ResNet56&ResNet110\\
Student&ResNet20&ResNet20&ResNet20&ResNet20\\
\midhline
T:Accuracy & 71.49 & 72.32 & 72.34 & 74.21 \\
S:Accuracy & 69.06 & 69.06 & 69.06 & 69.06  \\
\midhline
DKD & 71.23 & 71.74 & 71.97 & 70.91 \\
$\delta$ & {\color{red}-0.26} & {\color{red}-0.58} & {\color{red}-0.37} & {\color{red}-3.30} \\
\cline{1-5}
LDRLD & 71.45 & 72.00 & 72.20 & 74.16  \\
$\delta$ & {\color{red}-0.04} & {\color{red}-0.32} & {\color{red}-0.14} & {\color{red}-0.05} \\
\bottomline
\end{tabular}
}
\caption{The top-1 performance of students with different teachers on the CIFAR-100 datasets. $\delta$ represents the gap with the teacher's performance.}
\label{table:noisy-right}
\end{table}
\subsection{Combination with other KD}Table~\ref{combine kd} shows LDRLD is also compatible with other KD, with slight gains possibly due to redundant non-target knowledge transfer.
\begin{table}[!ht]
\centering
\renewcommand{\arraystretch}{1.4}
\resizebox{\linewidth}{!}{
\begin{tabular}{c|cccccc}
\tophline
Teacher & ResNet32$\times$4 & WRN-40-2 &ResNet32$\times$4 &VGG13 \\
Student&ResNet8$\times$4& WRN-40-1 & ShuffleNetV2 &VGG8  \\
\midhline
 LDRLD& 77.20&  74.98 & 77.33 &75.06 \\
 LDRLD+DKD& 77.42 (\textcolor{deepgreen}{+0.22})& 75.25 (\textcolor{deepgreen}{+0.27}) & 77.48 (\textcolor{deepgreen}{+0.15}) & 75.17 (\textcolor{deepgreen}{+0.11})   \\
 LDRLD+IKD~\cite{wang2025debiased}& 77.52(\textcolor{deepgreen}{+0.32})& 75.15 (\textcolor{deepgreen}{+0.17}) & 77.62 (\textcolor{deepgreen}{+0.29})  & 75.12 (\textcolor{deepgreen}{+0.06})   \\ 
\bottomline
\end{tabular}
}
\caption{Combining LDRLD with other KD on CIFAR100.}
\label{combine kd}
\end{table}
\section{Comparison with State-of-the-Art Methods}
\begin{table}[!ht]
\setlength{\belowcaptionskip}{-5pt} 
\center
\resizebox{\linewidth}{!}{
\begin{tabular}{c|ccc|ccc}
\tophline
\multicolumn{1}{c|}{ Manner} & \multicolumn{3}{c|}{R-101 \& R-18} & \multicolumn{3}{c}{R-101 \& R-50} \\
\tophline
    Metrics         & AP         & AP$_{50}$      & AP$_{75}$      & AP         & AP$_{50}$      & AP$_{75}$       \\ 
\midhline
Teacher               & 42.04      & 62.48     & 45.88     & 42.04      & 62.48     & 45.88       \\
Student               & 33.26      & 53.61     & 35.26     & 37.93      & 58.84     & 41.05       \\ 
\midhline
FitNet\cite{romero2014fitnets}                & 34.13      & 54.16     & 36.71     & 38.76      & 59.62     & 41.80       \\
FGFI~\cite{wang2019distilling}                 & 35.44      & 55.51     & 38.17     & 39.44      & 60.27     & 43.04       \\
ReviewKD\cite{chen2021distilling}              & 36.75      & 56.72     & 34.00     & 40.36      & 60.97     & 44.08       \\
\midhline
KD\cite{hinton2015distilling}                    & 33.97      & 54.66     & 36.62     & 38.35      & 59.41     & 41.71     \\
CTKD~\cite{li2023curriculum}&34.56& 55.43& 36.91&-&-&-\\
DKD~\cite{zhao2022decoupled}                   & 34.88      & 56.16     & 37.08     & 39.01      & 60.41     & 42.33     \\
LDRLD                 & \textbf{35.12}      & \textbf{56.79}     & \textbf{37.57}     & \textbf{39.31}      & \textbf{61.06}     & \textbf{42.56}     \\
\bottomline
\end{tabular}
}
\caption{Results on the MS-COCO using Faster-RCNN~\cite{ren2015faster}-FPN~\cite{lin2017feature}, with AP evaluated on val2017 dataset.}
\label{tab:coco}
\end{table}
\begin{table}[!ht]
\centering
\renewcommand{\arraystretch}{1.2} 
\resizebox{\linewidth}{!}{
\begin{tabular}{c|ccccc}
\tophline
Teacher  & ResNet32x4  & ResNet32x4     & VGG13       & VGG13 & ResNet50     \\ 
Accuracy      & 66.17       & 66.17          & 70.19       & 70.19 & 60.01        \\
\midhline
Student  & MobileNetV2 & ShuffleNetV1   & MobileNetV2 & VGG8  & ShuffleNetV1 \\ \
Accuracy      & 40.23       & 37.28          & 40.23       & 46.32 & 37.28        \\ \midhline
SP &48.49& 61.83& 44.28 &54.78 &55.31\\
CRD  &57.45  &62.28  &56.45  &66.10  &57.45\\
SemCKD  &56.89  &63.78  &68.23  &66.54  &57.20\\
ReviewKD &- &64.12& 58.66 &67.10 &-\\
MGD &-& - &- &66.89& 57.12\\
\midhline
KD       & 56.09       & 61.68          & 53.98       & 64.18 & 57.21        \\ 
DKD      & 59.94({\color{red}+3.85})       & 64.51({\color{red}+2.83})          & 58.45({\color{red}+4.47})       & 67.20({\color{red}+3.02}) & 59.21({\color{red}+2.00})        
     \\ 
NKD      & 59.81({\color{red}+3.72})       & 64.00({\color{red}+2.32})           & 58.40({\color{red}+3.42})       &    67.16({\color{red}+2.98})   &    59.11({\color{red}+1.90})         \\
LDRLD      &  \textbf{60.99}({\color{red}+4.90})       & \textbf{65.19}({\color{red}+3.51})           &  \textbf{59.73}({\color{red}+5.75})      &    \textbf{68.27}({\color{red}+4.09})   &    \textbf{60.46}({\color{red}+3.25})         \\ 
\bottomline
\end{tabular}
}
\caption{Top-1 accuracy (\%) of student on the CUB200 dataset.}
\label{fine-grained1}
\end{table}
\begin{table*}[!ht]
\begin{center}
\setlength\tabcolsep{8.0pt}
\scalebox{0.8}{
\begin{tabular}{c|c|ccccc}
\tophline
\multirow{2}*{Distillation} & Teacher &  ResNet56 & ResNet110 & VGG13 & WRN-40-2 & VGG13 \\
 & Accuracy & 56.56 & 59.01 & 60.20 & 60.45 & 60.20 \\
\cline{2-7}
\multirow{2}*{Manner} & Student & ResNet20 & ResNet20 & VGG8 & WRN-16-2 & MobileNetV2 \\
~ & Accuracy & 52.66 & 51.89 & 56.03 & 57.17 & 57.73 \\
\midhline
\multirow{11}{*}{Features} & AT~\cite{komodakis2017paying}& 54.39 & 54.57 & 58.85 & 59.39 & 60.84 \\
~ & SP~\cite{tung2019similarity} & 54.23 & 54.38 & 58.78 & 57.63 & 61.90 \\
~ & CC~\cite{peng2019correlation} & 54.22 & 54.26 & 58.18 & 58.83 & 61.32 \\
~ & VID~\cite{ahn2019variational} & 53.89 & 53.94 & 58.55 & 58.78 & 60.84 \\
~ & PKT~\cite{passalis2018learning} & 54.29 & 54.70 & 58.87 & 59.19 & 61.90 \\
~ & FT~\cite{kim2018paraphrasing} & 53.90 & 54.46 & 58.87 & 58.85 & 61.78 \\
~ & NST~\cite{huang2017like} & 53.66 & 53.82 & 58.85 & 59.07 & 60.59 \\
~ & FitNet~\cite{romero2014fitnets} & 54.43 & 54.04 & 58.33 & 58.88 & 61.37 \\
~ & RKD~\cite{park2019relational} & 53.95 & 53.88 & 58.58 & 59.31 & 61.19 \\
~ & CRD~\cite{tian2019contrastive} & 55.04 & 54.69 & 58.88 & 59.42 & 61.63 \\
~ & SAKD~\cite{song2022spot}& \textbf{55.06} & \textbf{55.28} & \textbf{59.53} & \textbf{59.87} & \textbf{62.29} \\
\midhline
\multirow{9}{*} & KD~\cite{hinton2015distilling} & 53.04 & 53.40 & 57.33 & 59.16 & 60.02 \\
& DKD~\cite{zhao2022decoupled}& 54.51 & 54.89 & 60.22 & 59.45 & 59.00 \\
{Logits}&WTTM~\cite{zheng2024knowledge}& 55.07* & 54.39* & \textbf{61.30*} & 59.95* & 60.06* \\
~&TeKAP~\cite{hossain2025single}& 54.83* & 54.58* & 60.37* & 59.66* & 59.08* \\
&LDRLD& \textbf{55.23} & \textbf{55.24} & 60.91 & \textbf{60.67} & \textbf{60.31} \\
& $\Delta$  & {\textbf{\color{deepgreen}+2.19}} & \textbf{{\color{deepgreen}+1.84}} & \textbf{{\color{deepgreen}+3.58}} & \textbf{{\color{deepgreen}+1.51}} & \textbf{{\color{deepgreen}+0.29}} \\
\bottomline
\end{tabular}}
\end{center}
\vspace{-0.6cm}
\caption{Evaluate the top-1 accuracy (\%) of students on the Tiny-ImageNet validation set.}
\label{table:tinyimagenet.}
\end{table*}
\subsection{Object Detection Results on MS-COCO2017}
We apply the LDRLD method to the object detection task and validate it on two different architectures: the teacher-student pairs are ResNet-101 (R-101) and ResNet-18 (R-18), as well as ResNet-101 and ResNet-50 (R-50). The experimental results show that our method outperforms DKD on the MS-COCO2017 dataset, demonstrating its good generalization ability.

\subsection{Fine-grained Task Results on the CUB200} The fine-grained task hinges on local features (e.g., feather patterns). LDRLD enhances inter-class separability by decoupling logits into combination pairs $[z_1,z_2]$, amplifying subtle differences while normalizing them to reduce interference from other categories. The results in Table~\ref{fine-grained1} show that LDRLD outperforms other KD and validates generalization ability (see the SDD training detail).
\subsection{Fine-grained Task Results on the Tiny-ImageNet.} To further validate the effectiveness of our method, we conducted experiments on Tiny-ImageNet, a fine-grained dataset with numerous categories and higher intra-class similarity.~These characteristics allow the teacher to effectively transfer fine-grained logit knowledge through LDRLD, thereby enhancing the student's performance.~As shown in Table~\ref{table:tinyimagenet.}, our method outperforms most existing classical KD approaches, achieving performance improvements ranging from 0.29\% to 3.58\% compared to KD. This achievement is primarily due to LDRLD enhancing the student's ability to capture detailed inter-class knowledge, which significantly boosts the student's discriminability in classification.

\section{Visualization.}
\begin{figure}[!ht]
    \centering
    \begin{subfigure}{0.23\textwidth}
        \centering
\includegraphics[width=\textwidth]{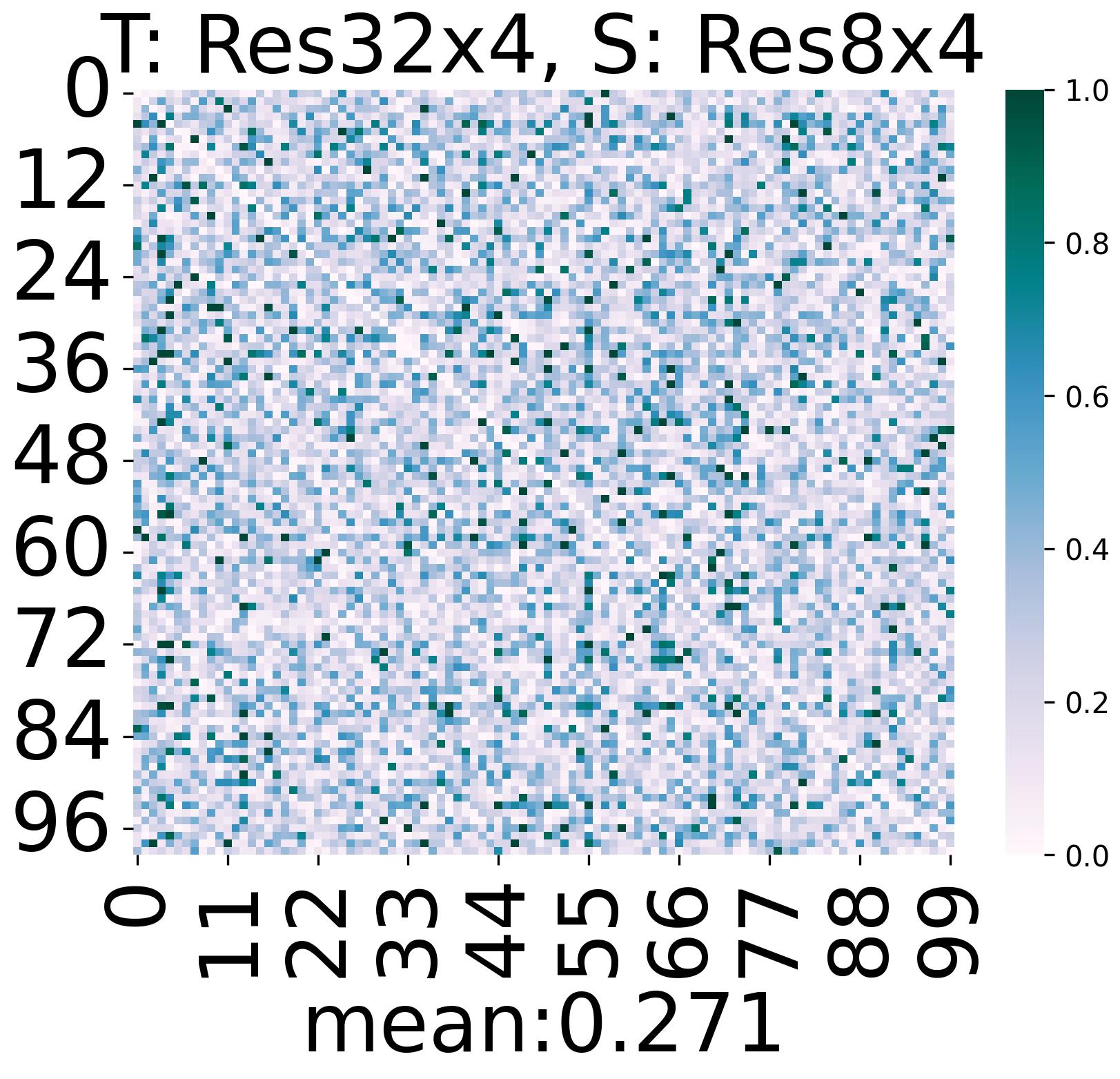}
\vspace{-12pt}
        \caption{KD}
    \end{subfigure}
    \begin{subfigure}{0.23\textwidth}
        \centering
\includegraphics[width=\textwidth]{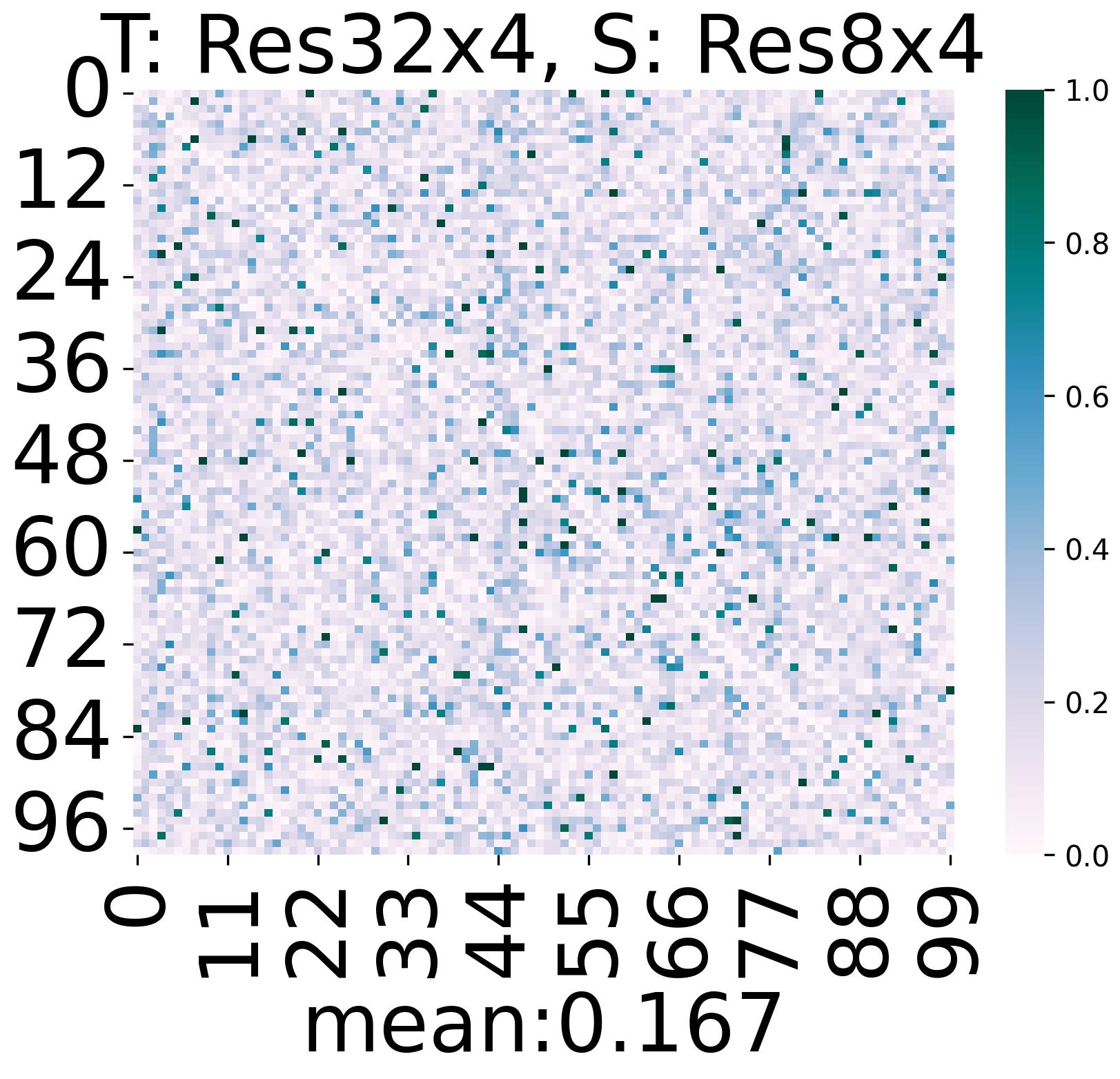}
\vspace{-12pt}
        \caption{LDRLD}
    \end{subfigure}
    \begin{subfigure}{0.23\textwidth}
        \centering
\includegraphics[width=\textwidth]{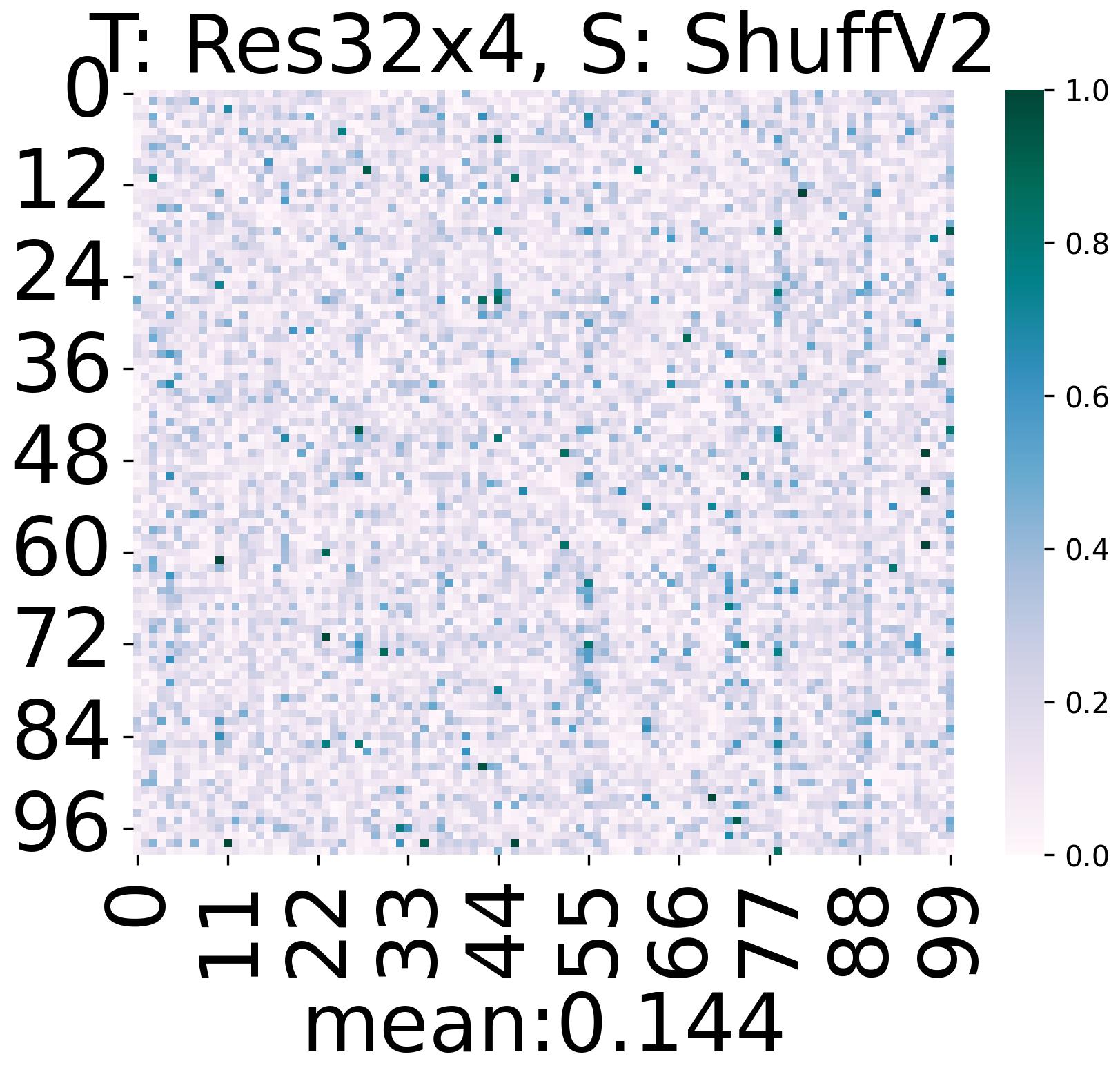}
\vspace{-12pt}
        \caption{KD}
    \end{subfigure}
    \begin{subfigure}{0.23\textwidth}
        \centering
\includegraphics[width=\textwidth]{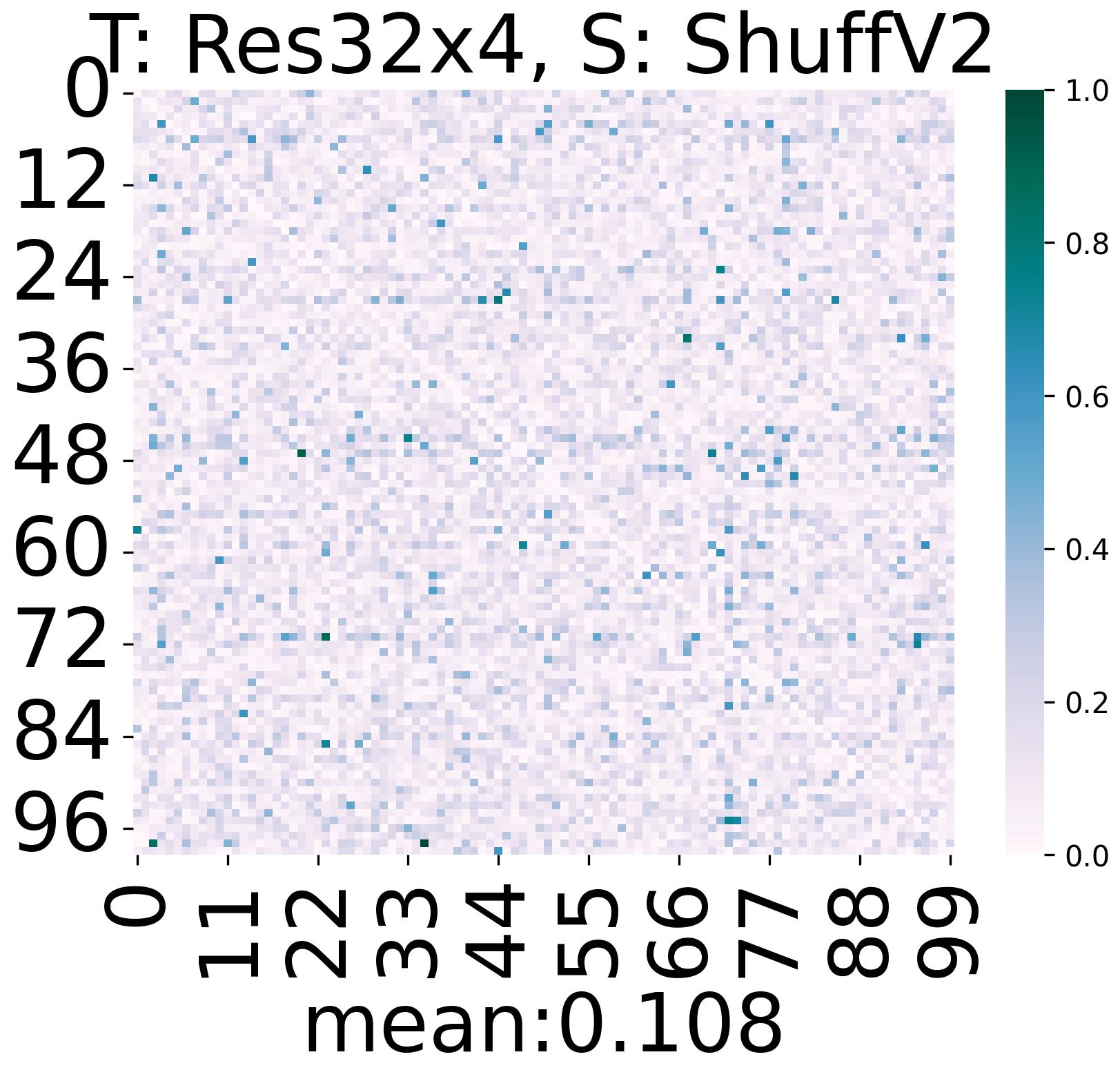}
\vspace{-12pt}
        \caption{LDRLD}
    \end{subfigure}
    \vspace{-8pt}
\caption{Visualization of the difference in correlation matrices between student and teacher logits for different teacher-student pairs: ResNet32$\times$4 vs ResNet8$\times$4, and ResNet32$\times$4 vs ShuffleNetV2, on the CIFAR-100 dataset.}
    \label{co-re}
\vspace{-13pt}
\end{figure}

\textbf{Visualization of Correlation Matrices.} By visualizing the difference in correlation matrices between the teacher and the student, we observe that LDRLD produces a smaller difference than vanilla KD, as shown in Figs.~\ref{co-re} (b) and (d) compared to Figs.~\ref{co-re} (a) and (c).~This finding indicates that the student's logits are closer to those of the teacher, confirming that our method facilitates more efficient knowledge acquisition. Consequently, it can be concluded that the enhanced knowledge transfer we propose can significantly improve the student's performance.
\vspace{-15pt}

\end{document}